\documentclass[11pt]{article}

\usepackage[margin=1in]{geometry}
\usepackage{setspace}
\usepackage{graphicx}
\usepackage{amsmath,amssymb,amsthm}
\usepackage{algorithm}
\usepackage{hyperref}
\usepackage{url}
\usepackage{indentfirst}
\usepackage{caption}
\usepackage{threeparttable}
\usepackage{booktabs}

\hypersetup{
    colorlinks=true,
    linkcolor=blue,
    citecolor=blue,
    urlcolor=blue
}

\title{\textbf{ICODEN: Ordinary Differential Equation Neural Networks for Interval-Censored Data}}

\author{
Haoling Wang\thanks{Department of Biostatistics and Health Data Science, University of Pittsburgh, PA, USA} 
\and
Lang Zeng\footnotemark[1]
\and
Tao Sun\thanks{Center for Applied Statistics and School of Statistics, Renmin University of China, Beijing, China}
\and
Youngjoo Cho\thanks{Department of Applied Statistics, Konkuk University, Seoul, Republic of Korea}
\and
Ying Ding\thanks{Department of Biostatistics and Health Data Science, University of Pittsburgh, PA, USA. Email: yingding@pitt.edu}
}

\date{}

\begin{document}

\maketitle

\begin{abstract}
Predicting time-to-event outcomes when event times are interval censored is challenging because the exact event time is unobserved. Many existing survival analysis approaches for interval-censored data rely on strong model assumptions or cannot handle high-dimensional predictors. We develop ICODEN, an ordinary differential equation-based neural network for interval-censored data that models the hazard function through deep neural networks and obtains the cumulative hazard by solving an ordinary differential equation. ICODEN does not require the proportional hazards assumption or a prespecified parametric form for the hazard function, thereby permitting flexible survival modeling. Across simulation settings with proportional or non-proportional hazards and both linear and nonlinear covariate effects, ICODEN consistently achieves satisfactory predictive accuracy and remains stable as the number of predictors increases. Applications to data from multiple phases of the Alzheimer's Disease Neuroimaging Initiative (ADNI) and to two Age-Related Eye Disease Studies (AREDS and AREDS2) for age-related macular degeneration (AMD) demonstrate ICODEN's robust prediction performance. In both applications, predicting time-to-AD or time-to-late AMD, ICODEN effectively uses hundreds to more than 1,000 SNPs and supports data-driven subgroup identification with differential progression risk profiles. These results establish ICODEN as a practical assumption-lean tool for prediction with interval-censored survival data in high-dimensional biomedical settings.

\noindent\textbf{Keywords:} Deep survival model; Interval censoring; Neural network; Ordinary differential equation
\end{abstract}

\section{Introduction}
\label{section-intro}

For progressive diseases such as dementia or Alzheimer's disease (AD), the precise time of disease onset or progression is often unobserved due to the intermittent nature of assessments. For example, a patient may be disease-free at one clinical visit and then diagnosed with the disease at the subsequent visit. As a result, these non-fatal events of interest (e.g., onset of AD) are only known to occur within an interval between two assessment times, producing interval-censored data \cite{klein2006survival}. Because interval-censored data frequently arise in clinical and epidemiologic studies, accurately modeling and predicting disease progression trajectories is of practical importance to understand disease evolution and guide clinical decision-making.

Right censoring is a special case of interval censoring in which the subsequent visit time is infinite and has been thoroughly investigated in the literature. Traditional survival analysis techniques for right-censored data, such as the Kaplan–Meier estimator and regression-based models, have been adapted to accommodate interval-censored observations and are implemented in the \textit{IcenReg} package \cite{icenReg}. This \textit{R} package includes the non-parametric Turnbull estimator, the semi-parametric Cox-proportional hazards (PH) model, and a selection of parametric models. In the era of big data, the presence of high-dimensional predictors, such as genome-wide SNP data and high-resolution imaging data, presents significant challenges for these approaches. Recently, several methods have been developed for interval-censored data with high-dimensional predictors, such as the Cox PH model with adaptive lasso \cite{li2020adaptive} and survival forest methods \cite{yao2021ensemble}. However, they cannot yet effectively handle thousands of SNPs in a dataset \cite{DUMMY:1(NN_IC)}.

A neural network (NN) is a flexible and powerful method for modeling the complex relationships inherent in high-dimensional data. It can take high-dimensional data directly as input and approximate intricate associations from data without strong assumptions on the form of the model \cite{bauer2019deep,schmidt2020nonparametric}. This flexibility allows NNs to learn complex, potentially nonlinear relationships among covariates.

Several NN-based survival analysis methods have been developed for right-censored data. DeepSurv, among the first NN survival approaches, replaces the Cox proportional hazards model's linear predictor with a neural network to capture complex covariate effects \cite{katzman2018deepsurv}. DeepHit and nnet-survival, two discretized-time survival NNs, do not rely on the PH assumption. Instead, they reformulate the continuous-time survival problem into a discrete-time problem, providing survival probabilities at predetermined times \cite{lee2018deephit,gensheimer2019scalable}. Recently, ordinary differential equation (ODE)-based NN frameworks have leveraged the intrinsic differential relationship between cumulative hazard and hazard functions, enabling continuous-time survival modeling through neural ODEs \cite{DUMMY:2(SODEN)}. SuMo-net further employs a monotone NN to directly model the survival function \cite{rindt2022survival}. Besides modeling the survival distribution, additional NN architectures have been applied for various medical applications, such as convolutional NNs and transformer-based architectures for clinical records and images with survival outcomes \cite{lv2022transsurv,zeng2025tdcoxsnn}. 

Recent studies have sought to extend these right‐censoring–based methods to accommodate interval‐censored observations. For example, BPNet adopts the PH assumption and models the baseline hazard function using Bernstein polynomials with a NN employed to learn the covariate effects \cite{DUMMY:1(NN_IC)}. There is also an accelerated failure time model-based NN method for interval-censored data \cite{meixide2024neural}. These frameworks, however, typically depend on strong assumptions regarding the form of the underlying survival distribution.

In this work, we propose a flexible continuous-time framework for analyzing high-dimensional, interval-censored survival data without imposing any specific structure (such as the PH assumption) on the underlying hazard or survival functions. Furthermore, to accommodate left truncation, which is frequently encountered in clinical trial settings, we extend the model to jointly handle both interval censoring and left truncation within the same ODE-based neural architecture.

The article is organized as follows. In Section 2, we present the architecture of our proposed ODE‐based NN for interval‐censored survival data. Section 3 examines a series of simulation studies with comparisons to alternative methods. In Section 4, we apply our method to multi-phase real-world research datasets of the Alzheimer's Disease Neuroimaging Initiative (ADNI) and two Age-Related Eye Disease Studies (AREDS and AREDS2), and compare its performance against established approaches. The conclusion and discussion are presented in Section 5.

\section{Methodology}
\label{section-method}

\subsection{Notation}
\label{subsection-notation}

In this section, we introduce the proposed ODE-based NN for the interval-censored data method, namely, ``ICODEN''. First, we introduce the notation for interval-censored data, which may be subject to left-truncation, as shown in our ADNI motivating example. Let $T_i$ be the (true) time to the event of interest, $V_i$ be the truncation time, and $X_i$ denote the covariate vector for the $i$-th individual. With left truncation, we can observe the $i$-th subject only if events occurred after a specific point in time ($V_i \leq T_i$) while subjects with $V_i>T_i$ are excluded from the study. In interval-censored survival data, the observed interval for the \( i \)-th individual is denoted as \( (L_i, R_i] \), indicating that the event occurred sometime after \( L_i \) and up to and including \( R_i \). For the individual whose event has not occurred by the last follow-up time \( L_i \), the observed interval is denoted as \( (L_i, +\infty) \) with $R_i=+\infty$. The observed data are i.i.d. samples $(V_i, L_i,R_i,X_i)$ with left truncation time $V_i$, and reduce to $(L_i,R_i,X_i)$ in the absence of left truncation.

The survival function $S\left( t | X \right)$ denotes the probability that the event time exceeds $t$, given the predictor $X$. We use $\lambda (t|X)$ and $ \Lambda (t|X) $ to denote the corresponding hazard function and cumulative hazard function at time $t$, respectively. The three functions are linked by 
$$S(t|X)=\exp (-\Lambda (t|X)) = \exp \bigg{(}-\int_0 ^t \lambda (s|X) \, \mathrm{d} s\bigg{)}. $$

\subsection{Likelihood for Interval-Censored Data}
\label{subsection-likelihood}

Next, we derive the full likelihood function for the interval-censored survival data, which will be used to optimize the ICODEN.  

Given the data structure discussed in the previous section, for the $i$-th individual, the probability that the event occurs within the observed interval is given by the difference in survival probabilities at the two endpoints, as shown in the following equation:

\begin{equation*}
    \mathrm{P} (T_i \in \left( L_i, R_i \right] | X_i) 
    = S(L_i | X_i) - S(R_i | X_i).
\end{equation*}

\noindent The full likelihood, expressed in terms of the survival function, is obtained by taking the product of these individual probabilities for all subjects. Alternatively, the likelihood can also be formulated using the cumulative hazard  
\begin{equation*}
\begin{aligned}
    L (L_i, R_i, X_i) 
    &= \prod_{i} \left(
    \exp \left(- \Lambda (L_i|X_i)  \right) 
    - \exp \left(-\Lambda (R_i|X_i) \right)
    \right) \text{.}
\end{aligned}
\end{equation*}
\noindent
By taking the logarithm of the likelihood function, we obtain the log-likelihood
\begin{equation}
\label{eq:llk}
    l (L_i, R_i, X_i) =\sum_i \log \left[
        \exp \left(- \Lambda (L_i|X_i)  \right) 
    - \exp \left(-\Lambda (R_i|X_i) \right)
    \right].
\end{equation}

When there is a left truncation at time $V_i$, the conditional probability for $T_i$ can be expressed as
$P(T_i \in \left( L_i, R_i \right] | T_i > V_i, X_i) =  \frac{S(L_i | X_i) - S(R_i | X_i)}{S(V_i|X_i)},$
and thus, the corresponding log-likelihood is
\begin{equation}
\label{eq:llk_LT}
    l_{LT} (V_i, L_i, R_i, X_i) =\sum_i \log \bigg{[}\frac{
        \exp \left(- \Lambda (L_i|X_i)  \right) 
    - \exp \left(-\Lambda (R_i|X_i) \right)
    }
    {\exp \left(- \Lambda (V_i|X_i)  \right)}\bigg{]}.
\end{equation}

\subsection{ICODEN Model}
\label{subsection-ode-nn-ic}

Instead of directly modeling the cumulative hazard function in Equation (\ref{eq:llk}), we propose using a NN to model the hazard function $\lambda(t|X)$ in Section \ref{subsubsec-NN part}, which does not have the monotonicity restriction. We then obtain the corresponding cumulative hazard function through an ordinary differential equation, as illustrated in Section \ref{subsubsec-ode part}.

\subsubsection{Neural Network Structure}
\label{subsubsec-NN part}

It is well known that NNs are universal approximators of continuous functions, allowing for flexible modeling of complex, non-linear relationships. They are also well-suited for handling high-dimensional input data, such as genetic information or other complex covariates.

\begin{figure}[h]
    \centering
    \includegraphics[width=0.5\textwidth]{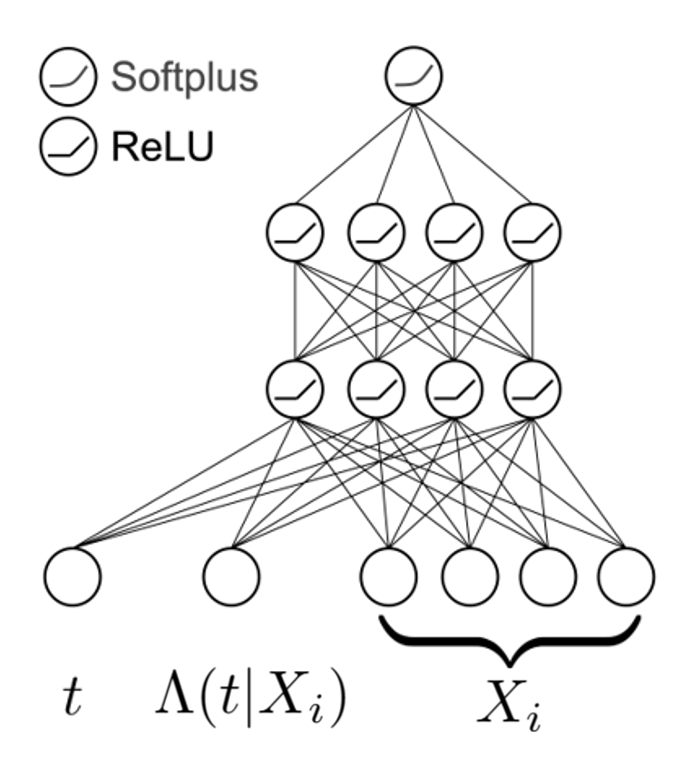}
    \caption{The proposed NN structure for $\lambda(t|X_i)$. The inputs are time $t$, predictor variables $X_i$, and the cumulative hazard function $\Lambda(t|X_i)$. Hidden layers use ReLU activation, while the output layer applies Softplus to ensure positivity.}
    \label{figure-NN structure}
\end{figure}

We propose a fully connected feedforward NN architecture for modeling the hazard function $\lambda(t|X_i)$. Specifically, the NN inputs are time $t$, predictor variables $X_i$, and the corresponding cumulative hazard function $\Lambda(t|X_i)$, allowing the hazard at time $t$ to depend on the previously accumulated hazard. This is consistent with the approach described in the previous work of Tang et al.~\cite{DUMMY:2(SODEN)}. The network consists of multiple hidden layers, each employing the ReLU activation function to introduce non-linearity. To ensure that the output, representing the hazard function, remains positive, the Softplus activation function $\text{Softplus}(x)=\ln(1+e^x)$ is applied to the output layer. We use $f_{\theta} (\Lambda(t|X_i), t, X_i)$ to denote the NN model for $\lambda(t|X_i)$, where $\theta$ is the parameter of the NN. The structure is shown in Figure \ref{figure-NN structure}.

\subsubsection{ODE for Cumulative Hazard Function}
\label{subsubsec-ode part}

In the proposed model $\lambda(t|X_i)=f_{\theta} (\Lambda(t|X_i), t, X_i)$, the cumulative hazard function $\Lambda(t|X_i)$ is included as one of the inputs to the network. As a result, the cumulative hazard must be computed iteratively along with the NN evaluation. This is achieved by using the relationship between the cumulative hazard and the hazard function, which can be formulated as an ODE in the following equation 
\begin{equation}
\label{eq: hazard-cumulative hazard relationship}
    \begin{cases}
        \frac{\mathrm{d}}{\mathrm{d}t} \Lambda(t|X_i, \theta) = \lambda(t|X_i, \theta) = f_{\theta} (\Lambda(t|X_i, \theta), t, X_i)\\
        \Lambda(0|X_i, \theta ) = 0.
    \end{cases}
\end{equation}
The second equation in \eqref{eq: hazard-cumulative hazard relationship} is the boundary condition at time $t=0$. Solving this ODE allows us to obtain the cumulative hazard values during the training process, which will be further introduced in Section 2.4. 

\subsection{Training the ICODEN}
\label{subsection-train tune}

The proposed model is trained by minimizing the negative log-likelihood introduced in (\ref{eq:llk}) with an additional $L_1$ regularization term to prevent the model from overfitting. This penalty term is controlled by a hyperparameter $\alpha$ to be tuned. Specifically, the fitted NN $f_\theta$ is the solution of the following optimization problem
\begin{equation*}
\label{eq:optimize}
    \arg \min_{\theta} \left\{-
     \sum_i \log \left[
        \exp \left( - \Lambda (L_i|X_i, \theta)  \right) 
    - \exp \left( - \Lambda (R_i|X_i, \theta) \right)
    \right]
    + \alpha \| \theta \|_1
    \right\},
\end{equation*}
subject to the ODE constraint in (\ref{eq: hazard-cumulative hazard relationship}).

We use the stochastic gradient descent-based optimization algorithm to optimize the target function.  To do so, with the current parameter value $\hat\theta^{(t)}$ at the $t$-th iteration step, the gradient of the log-likelihood with respect to the parameters $\theta$ in $\hat\theta^{(t)}$ is required. That is 
\begin{eqnarray*}
       && \nabla_{\theta} l(\theta)\rvert_{\theta = \hat\theta^{(t)}}= 
       \quad  \sum_i
    \frac{
    e^{ - \Lambda (R_i|X_i, \hat\theta^{(t)}) } \frac{\partial}{\partial \theta} \Lambda (R_i|X_i, \theta)\rvert_{\theta = \hat\theta^{(t)}}
    -
    e^{- \Lambda (L_i|X_i, \hat\theta^{(t)})} \frac{\partial}{\partial \theta} \Lambda (L_i|X_i, \theta)\rvert_{\theta = \hat\theta^{(t)}} 
    }
    {e^{ - \Lambda (L_i|X_i, \hat\theta^{(t)})} 
    - e^{ - \Lambda (R_i|X_i, \hat\theta^{(t)}) }}.
\end{eqnarray*}
Notice that $\Lambda (R_i|X_i, \hat\theta^{(t)})$ is the function value of $\Lambda (t|X_i, \hat\theta^{(t)})$ at $t=R_i$, where $\Lambda (t|X_i, \hat\theta^{(t)})$ is the solution of the ODE
\begin{equation}
\label{eq: gradient of loss function}
    \begin{cases}
        \frac{\mathrm{d}}{\mathrm{d}t} \Lambda(t|X_i, \hat\theta^{(t)})  = f_{\hat\theta^{(t)}} (\Lambda(t|X_i, \hat\theta^{(t)}), t, X_i),\\
        \Lambda(0|X_i, \hat\theta^{(t)} ) = 0.
    \end{cases}
\end{equation}
Therefore, we can solve the above ODE and plug in $t=R_i$ for $\Lambda (R_i|X_i, \hat\theta^{(t)})$. This step has been well implemented in an existing Python ODE solver \textit{torchdiffeq} \cite{torchdiffeq}, which not only computes $\Lambda(R_i | X_i, \hat\theta^{(t)})$ numerically but also provides $\frac{\partial}{\partial_\theta}\Lambda(R_i | X_i, \theta)\rvert_{\theta=\hat\theta^{(t)}}$ simultaneously. Similarly, we can obtain $\Lambda(L_i | X_i, \hat\theta^{(t)})$ and $\frac{\partial}{\partial \theta}\Lambda(L_i |X_i, \theta)\big\rvert_{\theta=\hat\theta^{(t)}}$. These quantities enable the computation of $\nabla_{\theta} l(\theta) $, which is then used to update the optimization algorithm for $\hat\theta^{(t+1)}$.

The ODE in (\ref{eq: gradient of loss function}) is individual-specific, and solving it separately for each individual is computationally expensive. We adopt the time-rescaling trick in \cite{DUMMY:2(SODEN)} to accelerate computation by exploiting GPU-based implementation of ODE solvers. Details of the trick are described in Section S3 of the Supplementary Material.

An early stopping rule is implemented during training to prevent overfitting and save computing resources. Specifically, if the validation loss does not decrease in 10 consecutive epochs, the training process is terminated. The model is considered well-trained at this point and is selected as the final output.

To optimize NN performance, we carefully tuned a set of hyperparameters through cross-validation. The hyperparameters include the number of nodes in each hidden layer, the number of training epochs, the $L_1$ regularization parameter, the batch size, and the learning rate. The tuning procedure and the final set of selected hyperparameters used in the model are introduced separately in Section S2 of the Supplementary Material.

\subsection{Evaluation Metrics}
\label{subsection:evaluation}

In simulation studies, the true survival function $S(t|X_i)$ and the true event time $T_i$ are known. We use the mean squared prediction error (MSE), which is essentially the average integrated $L_2$ distance between the true and estimated survival functions, 
\begin{equation*}
\label{eq: MSE}
    MSE=\frac{1}{n} \sum_{i=1}^{n} \frac{1}{T_i} 
                            \int_{0}^{T_i} {\left\{ 
                            S(t|X_i) - \hat{S} (t|X_i)
                            \right\}}^2 \, \mathrm{d}t.
\end{equation*}
A smaller MSE indicates better model performance.

In the real data analysis, the true survival time and the corresponding survival function are unknown. Instead, only the time interval $(L_i, R_i]$  is available. As a result, direct evaluation of the estimated survival function is not feasible. Instead, we propose using the following surrogate quantity for the true survival distribution, as defined in (\ref{eq: surrogate survival}), to compute a metric comparable to the MSE.
\begin{equation}
\label{eq: surrogate survival}
    \hat{I}(T_i>t | X_i) = 
    \begin{cases}
        1 , & t \leq L_i \\
        \frac{\hat{S}(t|X_i)-\hat{S}(R_i|X_i)}{\hat{S}(L_i|X_i)-\hat{S}(R_i|X_i)}, & L_i < t \leq R_i\\
        0 , & t > R_i
    \end{cases}
\end{equation}
This leads to the evaluation metric integrated Brier Score (IBS) \cite{IBS}, defined as:
\begin{equation*}
    IBS=\frac{1}{n} \sum_{i=1}^{n} \frac{1}{u} 
                            \int_{0}^{u} {\left\{ 
                            \hat{I}(T_i>t | X_i) - \hat{S} (t|X_i)
                            \right\}}^2 \, \mathrm{d}t,
\end{equation*}
where $u$ is the maximum finite value of all observed $L_i$ and $R_i$.

In the real data analysis, two additional evaluation metrics are utilized to further assess model performance under interval-censoring. The first metric, denoted as $p_{\mathrm{out}} = \mathrm{P}(\hat{t}_i^{0.5} \notin (L_i, R_i])$, is the proportion of predicted median survival times $\hat{t}_i^{{0.5}}$ that fall outside the observed censoring interval. The second metric, referred to as $d_{\mathrm{out}} =\min \{|L_i-\hat{t}_i^{0.5}|, |R_i-\hat{t}_i^{0.5}|\}$, measures the absolute distance between the predicted median survival time and the closest interval boundary. 

\section{Simulation}
\label{section-simulation}

\subsection{A Simple Example with One Binary Predictor}
\label{subsection: simple example}

We present a simple example with a single binary predictor to demonstrate the effectiveness of our proposed method. 

The predictor $X_i$ is generated from the Bernoulli distribution with probability $0.5$. The event time is generated from a mixture of two distributions that do not satisfy the PH assumption: 
\begin{equation*}
    S(t|X_i) = \exp (-2t^2) \, \mathrm{I}(X_i=1)+\exp (-2t) \, \mathrm{I}(X_i=0).
\end{equation*}
We simulated interval-censored data by generating 20 independent random intervals and recording the interval that covers the true event time. Specifically, for each sample, the length of each interval $\Delta T$ is drawn from an exponential distribution with a rate of 0.1.  The endpoints of the intervals, which represent the timing of clinic visits, are determined by the cumulative sum of these interval lengths. The observed interval $(L_i, R_i)$ for each individual is defined as the interval that contains the true event time, thereby representing the interval-censored nature of the data. We simulated $N=1,000$ observations and fit the model based on observed intervals $\{(L_i, R_i)\}_{i=1}^{N}$. 

\subsection{Multiple Predictors}
\label{subsection:gen case}

We further conducted a set of simulations with multiple predictors. We considered four scenarios (presented below), each with $p=20, 50$ and $300$ predictors, representing varying levels of complexity. These scenarios are used to assess the scalability and robustness of the models in higher-dimensional covariate spaces. The predictor variables for $i$-th individual $X_i$ are generated from a multivariate normal distribution of $MVN(0,\Sigma)$, where the covariance matrix is $\Sigma=\{\sigma_{jj'}=e^{-|j-j'|} ,\, 1\leq j,j' \leq p\}$. This covariance matrix assumes the correlation between variables. The first $ 20 \%$ variables remain continuous. The second $ 20 \% $ of the variables are transformed to binary predictors following the indicator function $I(X_{ij}>0)$ and the remaining $ 60 \% $ predictors are transformed into multinomial predictors by the sum of two indicator functions $I(X_{ij}>-0.5)+I(X_{ij}>0.5)$. The discretized variables are used to mimic the alleles in the SNP data. 

After generating the covariates, we simulated event times from a range of survival functions encompassing both proportional hazards and non-proportional hazards settings, as well as linear and non-linear relationships between predictors and the log hazard function. This allows for an assessment of model flexibility and robustness in various functional forms. For the PH model (Scenarios 1 and 2), the event time $T_i$ for each individual has a baseline survival from a Weibull distribution with $S_0(t)=e^{-0.01t^{10}}$. For the non-PH model, the baseline hazard is determined by a single covariate, as shown in Scenarios 3 and 4 below. The effect of the remaining predictors is incorporated through either a linear or a non-linear functional form in the survival function. The coefficients $\beta_i$ for continuous and binary predictors are set to 0.2, while the coefficients for multinomial predictors are generated from $MVN (0.2, 0.01\times\Sigma)$ (and are held constant throughout the simulation runs within each scenario). 

\begin{algorithm}[ht]
\caption*{\textbf{Simulation Scenarios}}
\label{eq:scenarios}
     \textbf{S1 (PH + linear effect):
     } 
     $S (t|X_i) = e^{-0.01t^{10} \exp \left(
            \sum_{j=1}^{p} \beta_j X_{ij}
            \right)}$
            
     \textbf{S2 (PH + nonlinear effect): 
     }
    $S(t|X_i) = e^{-0.01t^{10} \exp \left(
            \sum_{j=1}^{p} \beta_j X_{ij} + X_{i1}^2 + X_{i2}^2 + X_{i3}X_{i4}
            \right)}$

     \textbf{S3 (non-PH +linear effect): 
     }
    $S(t|X_i)=
        \begin{cases}
            e^{-0.01t^{10} \exp \left(
            \sum_{j=1}^{p} \beta_j X_{ij}
            \right)} , & (X_{i,b} = 0)
            \\
            e^{-0.01t^{5} \exp \left(
            \sum_{j=1}^{p} \beta_j X_{ij}
            \right)} , & (X_{i,b} = 1)
        \end{cases}$

    \textbf{S4 (non-PH + nonlinear effect): 
     }\\
     $S(t|X_i)=
    \begin{cases}
        e^{-0.01t^{10} \exp \left(
        \sum_{j=1}^{p} \beta_j X_{ij} + X_{i1}^2 + X_{i2}^2 + X_{i3}X_{i4}
        \right)} , & (X_{i,b} = 0)
        \\
        e^{-0.01t^{5} \exp \left(
        \sum_{j=1}^{p} \beta_j X_{ij} + X_{i1}^2 + X_{i2}^2 + X_{i3}X_{i4}
        \right)} , & (X_{i,b} = 1)
    \end{cases}$

($X_{i,b}$ in S3 and S4 denotes the first binary covariate.)
\end{algorithm}

To reflect variability in the number of clinical visits among individuals, we simulated the number of observation intervals for each subject from a Poisson distribution with a mean of 12. Based on the generated number of visits, we then constructed observation intervals following the same procedure used in the simple example simulation, thereby generating individual-specific interval-censored data. 

We generated 1,000 samples for the training set and another 1,000 samples as the held-out test set to evaluate model performance. Each simulation scenario is repeated 100 times. The proposed method is compared with the semi-parametric Cox-PH and the ALASSO (adaptive LASSO) methods for interval-censored data, implemented in the R packages \textit{IcenReg} and \textit{ALassoSurvIC}. The model performance is evaluated using the MSE of the predicted survival probabilities.

\subsection{Synthetic Simulation}
\label{subsection:synthetic}

We also employed a synthetic data generation approach that retains real-world covariates while simulating survival times and censoring indicators based on a known survival function or risk model. This approach enables controlled evaluation of survival models under realistic covariate structures.

Specifically, we randomly selected 1000 individuals from the ADNI dataset as training samples, and the remaining individuals as test samples. To generate the true survival time, we used variables of age, education year, gender, whether they have an \textit{APOE} allele, and 200 other randomly selected SNPs. The true survival function used to simulate the event time is presented below, which contains a non-proportional hazards component determined by the \textit{APOE} allele and an additional component capturing both linear and non-linear effects of the selected predictors. The observed interval is generated similarly to the method described in Section \ref{subsection:gen case}. The subscript $j$ for the predictors indexes the randomly selected 200 SNPs, and $a$, $e$, and $g$ are scaled age, scaled education year, and gender, respectively. The coefficients $\beta_j$ were generated from $MVN (0.2, 0.01\times\Sigma)$ and held constant throughout all simulations. 

\begin{algorithm}[ht]
\caption*{\textbf{Synthetic Simulation}}
\begin{equation*}
\label{eq: synthetic simulation}
S(t|X_i)=
    \begin{cases}
        e^{-0.01t^{10} \exp \left(
        \sum_{j} \beta_j X_{ij} + X_{i, a}^2 + X_{i,g}X_{i,e}
        \right)} , & (APOE = 1,2)
        \\
        e^{-0.01t^{5} \exp \left(
        \sum_{j} \beta_j X_{ij} + X_{i, a}^2 + X_{i,g}X_{i,e}
        \right)} , & (APOE = 0)
    \end{cases}
\end{equation*}
\end{algorithm}

During the model fitting, we also included 171 noise SNPs in addition to the 200 signal SNPs. This setting allows for assessing the model's ability to identify relevant signals in the presence of high-dimensional noise.

\subsection{Simulation Results}
\label{subsection: simu result}

\begin{figure}[ht]
    \centering
    \includegraphics[width=0.48\textwidth]{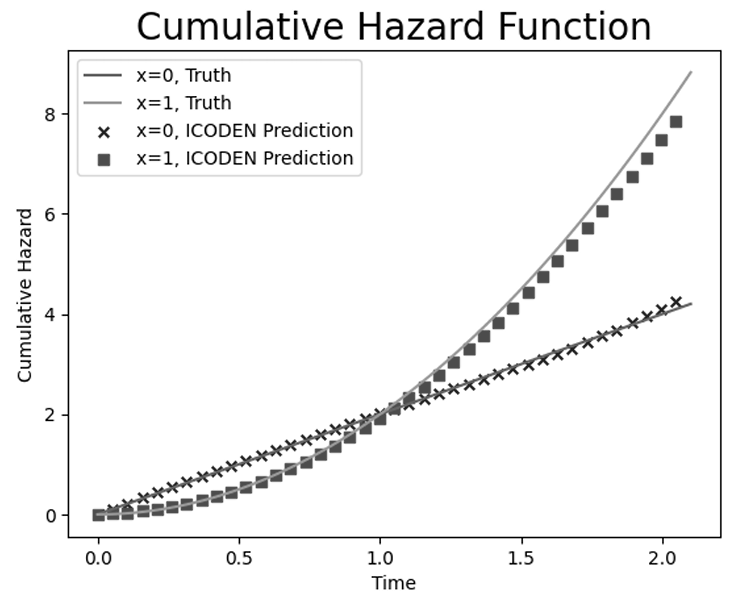}
    \includegraphics[width=0.48\textwidth]{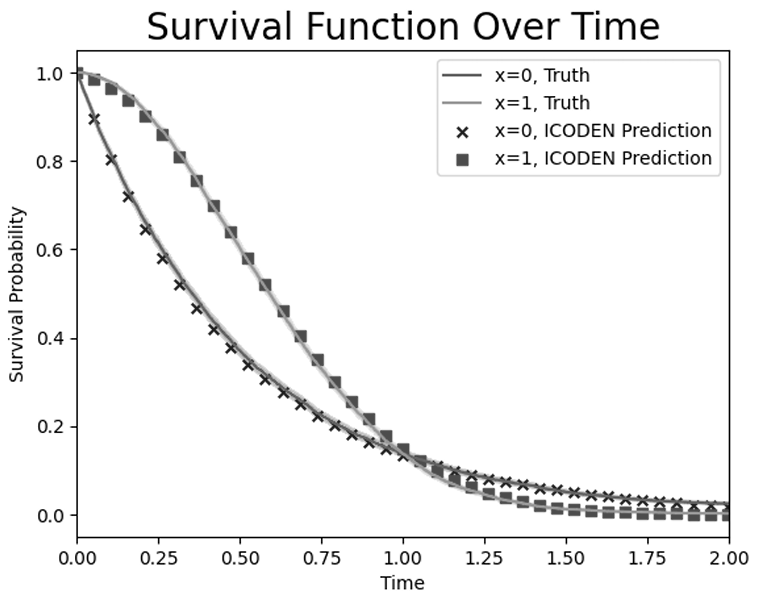}
    \caption{Simulation with a binary predictor under interval censoring and proportional hazards violation. 
    The left panel shows the fitted cumulative hazard functions. The right panel shows the corresponding fitted survival functions.}
    \label{simulation-example result}
\end{figure}
 
The simulation results for section \ref{subsection:gen case} are presented in Table \ref{table: Simulation Result}. In Scenarios 1 and 2, where the PH assumption holds, IcenReg and ALASSO generally produce better predictive performance, except when the number of predictors is 300 and the effects are non-linear. This is expected because the data-generating mechanisms in Scenarios 1 and 2 align with the assumption in IcenReg and ALASSO, giving them a modeling advantage over ICODEN in these cases. In Scenarios 3 and 4, where the PH assumption is violated, the ICODEN consistently outperforms IcenReg and ALASSO. Overall, the proposed ICODEN model demonstrates superior performance in capturing complex data structures, including non-PH and non-linear predictors. Note that in high-dimensional non-PH settings (Scenarios 3 and 4 with $p=300$), ALASSO did not converge and therefore produced no estimates. 

For the synthetic scenario described in Section \ref{subsection:synthetic}, as shown in the last row of Table \ref{table: Simulation Result}, ICODEN achieves the best predictive accuracy and even outperforms Scenario 4 with $p=300$, which has a very similar true survival functional form and the number of predictors as this synthetic scenario. This suggests that ICODEN is better suited to the high-dimensional and noisy structure of the real-world datasets that we will analyze.

\begin{table}[ht]
\centering
\caption{Simulation results for the multiple-predictor and synthetic settings. Values are the MSE reported as mean (SD). ICODEN is the proposed method; IcenReg refers to the Cox-PH model; ALASSO refers to the adaptive LASSO model.}
\label{table: Simulation Result}
\begin{tabular}{lllll}
\toprule
 & Predictors & ICODEN & IcenReg & ALASSO\\ 
\midrule
 
Scenario 1 & 20 & 0.018 (0.005) & 0.001 (0.0001) & 0.006 (0.001)\\
(PH, linear) & 50 & 0.048 (0.083) & 0.002 (0.0004) & 0.010 (0.001)\\ 
 & 300 & 0.115 (0.014) & 0.049 (0.003) & 0.074 (0.004) \\
\midrule

Scenario 2 & 20 & 0.034 (0.009) & 0.020 (0.001) & 0.024 (0.002)\\
(PH, non-linear) & 50 & 0.034 (0.014) & 0.023 (0.001) & 0.026 (0.001)\\ 
 & 300 & 0.155 (0.010) & 0.234 (0.014) & 0.204 (0.011) \\
\midrule

Scenario 3 & 20 & 0.015 (0.002) & 0.008 (0.001) & 0.021 (0.001)\\
(Non-PH, linear) & 50 & 0.108 (0.006) & 0.144 (0.006) & 0.130 (0.005)\\ 
 & 300 & 0.216 (0.008) & 0.435 (0.012) & -- \\
\midrule

Scenario 4 & 20 & 0.046 (0.003) & 0.064 (0.004) & 0.069 (0.004)\\
(Non-PH, non-linear) & 50 & 0.187 (0.007)  &  0.211 (0.007) & 0.193 (0.006)\\ 
 & 300 & 0.207 (0.007) & 0.358 (0.032) & --\\
\midrule

Synthetic Scenario&   &   &   \\
(Non-PH, non-linear) & 375 & 0.168 (0.006) & 0.349 (0.009) & 0.270 (0.007)  \\
\bottomrule
\end{tabular}
\end{table}

\section{Real Data Analysis}
\label{section-realData}

We applied our ICODEN model to two datasets with interval-censored outcomes. The first is the ADNI dataset to predict time-to-AD progression, and the second is the AREDS \& AREDS2 datasets to predict time-to-late AMD. Both diseases are chronic degenerative disorders. In addition to comparing model predictions, we also identified subgroups using the predicted cumulative hazard at a pre-specified time for each individual. This allows us to discover distinct risk groups for disease progression.

\subsection{Analysis of AD Progression in ADNI Data}
\label{subsection:ADNI}

ADNI is a multi-phase, longitudinal, and multi-center observational study aimed at developing clinical, imaging, and genetic biomarkers for the early detection and monitoring of AD \cite{mueller2005alzheimer}. ADNI has recruited over 2,300 participants, including individuals with cognitively normal (CN) status, mild cognitive impairment (MCI), and AD, who were intermittently assessed during follow-up. The study consists of five phases: ADNI-1 (2004–2010), ADNI-GO (2009–2011), ADNI-2 (2011–2016), ADNI-3 (2016–2022), and ADNI-4 (2022–present). In this work, we use data collected from the first four phases, ADNI-1, ADNI-GO, ADNI-2, and ADNI-3. All four phases contain interval-censored data, and ADNI-GO also has left truncation since subjects with AD are not included in the phase of ADNI-GO. The analysis data used in our study were downloaded on May 1, 2021.

We analyzed 1,740 Caucasian participants from four ADNI phases with baseline demographics including age, gender, and education years, along with \textit{APOE} allele status and a list of other SNPs, which came from a previous genome-wide association study (GWAS) that identified SNPs associated with AD \cite{DUMMY:1(NN_IC)}. We selected three SNP lists using significance thresholds of $p=1\times 10^{-5}$, $1\times 10^{-4}$, and $2\times 10^{-4}$, respectively, to examine model performance under different numbers of predictors. These thresholds yielded 71, 371, and 623 SNPs, respectively. In this study, age 55 was used as the time origin (for the progression of time-to-AD), as the ADNI database only includes participants aged 55 and older, and AD is rare in individuals younger than 55 \cite{reitz2020late}. The characteristics of the analysis data are shown in the table \ref{table: descriptive}. In this analysis dataset, 63.1\% of individuals did not develop AD by the end of follow-up. 

\begin{table}[ht]
\centering
\caption{Descriptive characteristics of the ADNI dataset}
\label{table: descriptive}
\begin{tabular}{lc}
\toprule
 \textbf{Variable} & N=1740\\ 
\midrule
 \textbf{Age (Years)} & \\
 Mean (SD) & 73.5 (7.1) \\
 Median (Range) & 73.5 (55.0-91.4) \\ 
\midrule
 \textbf{Gender (n, \%)} & \\
 Female & 780 (44.8\%) \\
 Male & 960 (55.2\%) \\ 
\midrule
 \textbf{Education (Year)} & \\
 Mean (SD) & 16.1 (2.8) \\
 Median (Range) & 16 (4-20) \\ 
\midrule
 \textbf{\textit{APOE} Allele (n, \%)} & \\
 0 Allele & 941 (54.1\%) \\
 1 Allele & 637 (36.6\%) \\
 2 Allele & 162 (9.3\%) \\ 
\midrule
 \textbf{Censoring Type (n, \%)} & \\
 Left-censored & 300 (17.2\%) \\
 Interval-censored & 342 (19.7\%)\\
 Right-censored & 1098 (63.1\%)\\ 
\midrule
 \textbf{ADNI Phase (n, \%)} & \\
 ADNI-1 & 703 (40.4\%) \\
 ADNI-GO & 114 (6.5\%) \\
 ADNI-2 & 619 (35.6\%) \\
 ADNI-3 & 304 (17.5\%) \\ 
\bottomrule
\end{tabular}
\end{table}

\subsubsection{Prediction Model Comparisons}
\label{subsubsec:ADNI models}

For each selected GWAS significance threshold, we performed five-fold cross-validation on the ADNI dataset using the corresponding SNPs below the threshold along with other baseline characteristics. We compared the performance of three different models: our proposed model (``ICODEN''), an interval-censored Cox PH model with only \textit{APOE}, gender, and education (``APOE''), and a similar interval-censored Cox regression model with individual SNPs in addition to gender and education (``IcenReg''). Both ``APOE'' and ``IcenReg'' models were implemented using the \texttt{IcenReg} package in R \cite{icenReg}. The ALASSO model was not included here because both our simulation study and other prior analyses on the ADNI data \cite{DUMMY:1(NN_IC)} found that this method cannot work stably when the number of variables approaches 300. After fitting the models, we evaluated their performance on the held-out test data using the IBS score, $p_{\mathrm{out}}$ and $d_{\mathrm{out}}$. 

The results are shown in Table \ref{table: ADNI model compare}. We observe that the APOE model performs the worst among the three models, highlighting the benefit of including additional SNPs to enhance predictive accuracy. Based on the evaluation metrics $p_{out}$ and $d_{out}$, the performance of ICODEN is generally comparable to that of the IcenReg model, particularly when the predictor number is small (71 and 371 SNPs). In terms of the IBS score, ICODEN performs comparably to IcenReg when using 71 SNPs and outperforms IcenReg when using 371 SNPs. Once the number of SNPs increased to 623, the IcenReg model failed to run, leaving the ICODEN model as the only approach capable of handling this level of high-dimensional predictors.

\begin{table}[ht]
\centering
\begin{threeparttable}
\caption{Performance comparison of various models using ADNI data. Values are reported as mean (SD). ICODEN denotes the proposed method; APOE and IcenReg are Cox–PH models using only the \textit{APOE} gene and all SNPs, respectively.}
\label{table: ADNI model compare}
\begin{tabular}{llllll}
\toprule
& P-Value\tnote{a} & SNPs & APOE\tnote{b} & IcenReg & ICODEN \\ 
\midrule

IBS & 
$<1\times 10^{-5}$ & 71 & 0.081 (0.00004) & 0.054 (0.0005) & 0.061 (0.004) \\
& $<1\times 10^{-4}$ & 371 & 0.081 (0.00004) & 0.065 (0.002) & 0.060 (0.002) \\
& $<2\times 10^{-4}$ & 623 & 0.081 (0.00004) & -- & 0.068 (0.007) \\ 
\midrule

$p_{out}$ & 
$<1\times 10^{-5}$ & 71 & 0.40 (0.002) & 0.34 (0.005) & 0.35 (0.003) \\
& $<1\times 10^{-4}$ & 371 & 0.40 (0.002) & 0.34 (0.002) & 0.32 (0.004) \\
& $<2\times 10^{-4}$ & 623 & 0.40 (0.002) & -- & 0.32 (0.017) \\ 
\midrule

$d_{out}$ & 
$<1\times 10^{-5}$ & 71 & 3.79 (0.01) & 2.58 (0.03) &  2.80 (0.15) \\
& $<1\times 10^{-4}$ & 371 & 3.79 (0.01) & 3.02 (0.11) & 2.99 (0.08) \\
& $<2\times 10^{-4}$ & 623 & 3.79 (0.01) & -- & 3.53 (0.19) \\ 
\bottomrule
\end{tabular}
\begin{tablenotes}[flushleft]\footnotesize
    \item \tnote{a} GWAS P-value threshold from a previous study \cite{DUMMY:1(NN_IC)}.
    \item \tnote{b} The APOE model is invariant to the choice of p-value thresholds. 
\end{tablenotes}
\end{threeparttable}
\end{table}

\subsubsection{Subgroup Identification}
\label{subsubsec:subgroup ADNI}

\begin{figure}[ht]
    \centering
    \includegraphics[width=0.7\textwidth]{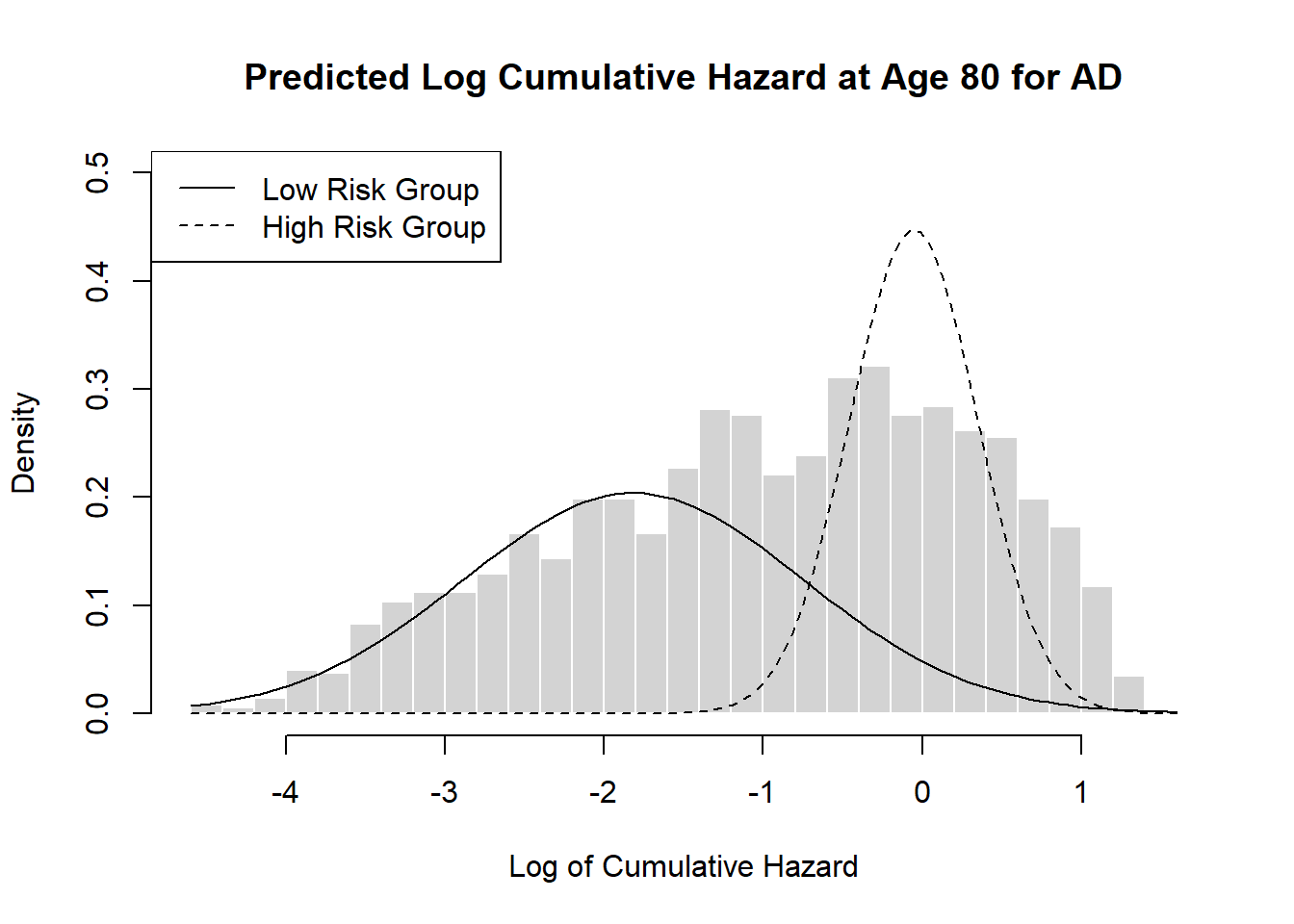}
    \includegraphics[width=0.7\textwidth]{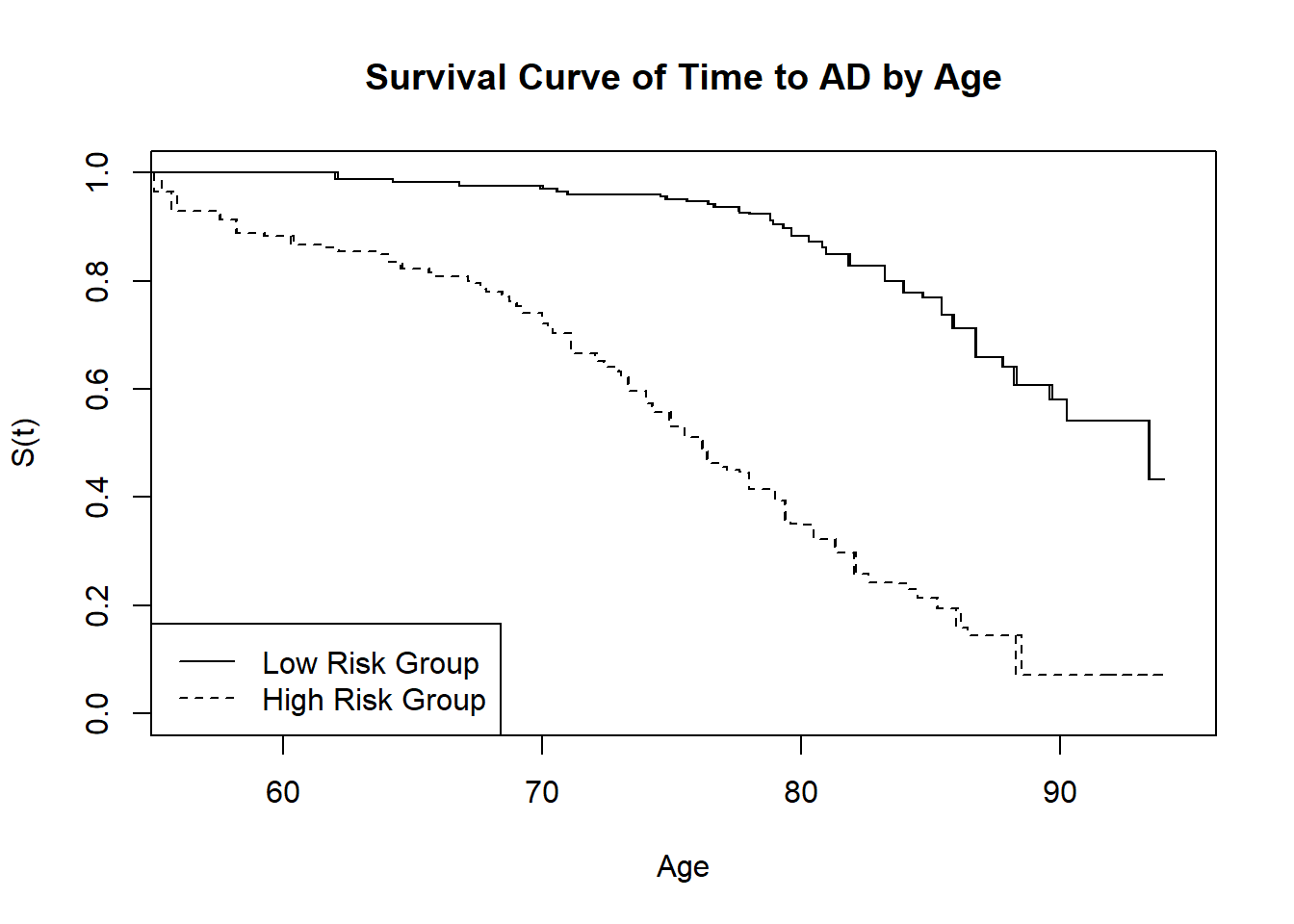}
    \caption{Gaussian mixture model–based subgrouping of ADNI participants. The left panel shows the density distribution of predicted log cumulative hazard at age 80, with the fitted Gaussian components corresponding to low- and high-risk groups. The right panel shows the Turnbull survival estimates for age at AD onset in the two subgroups.}
    \label{ADNI gaussian mixture model}
\end{figure}

We applied a Gaussian mixture model to the predicted log-cumulative hazards at age 80 of patients in the test dataset and identified two subgroups. Figure \ref{ADNI gaussian mixture model} presents the distribution of the predicted log-cumulative hazards along with the Turnbull estimator for the survival probability of each subgroup. Since exact event times are not observed under interval censoring, the Turnbull estimator yields a survival band rather than a single curve, in contrast to the Kaplan–Meier estimator for right-censored data \cite{turnbull1976empirical,icenReg}. 

The right panel of Figure \ref{ADNI gaussian mixture model} clearly shows that the two subgroups exhibit significantly different estimated survival functions (log-rank test $P<2.2\times 10^{-16}$), demonstrating that the proposed method can successfully identify subgroups with distinct risk profiles. Importantly, the identified subgroups are biologically meaningful. When comparing the distribution of \textit{APOE} allele $rs429358$ between the two groups, we observed a significant difference (Chi-square test $P<4.82 \times 10^{-58}$; see Supplemental Table S4.1), consistent with the well-known \textit{APOE} effect on AD risk. Other SNPs are also examined in the two identified groups. The top 10 SNPs with the smallest P-value between the two identified groups are also shown in the Supplemental Table S4.1. Notably, another well-known SNP within the \textit{APOE} gene, $rs7412$, was among the top signals, further supporting consistency with previous findings \cite{zuo2006variation}.

\subsection{Analysis of AMD Progression in AREDS and AREDS2 Data}
\label{subsection:AREDS}

AREDS and the following AREDS2 are two multicenter, controlled, randomized clinical trials of AMD and age-related cataract, sponsored by the National Eye Institute. It was designed to assess the clinical course and risk factors for the development and progression of AMD and cataract. The study uses a severity score from 1 to 12 (a higher score indicates more severe AMD) to measure disease progression. It also collected DNA samples of consenting participants and performed genome-wide genotyping \cite{areds,areds2}. 

In this analysis, we analyzed 7803 eyes from 4394 participants in AREDS and AREDS2, which were at the early to intermediate stage (defined as a severity score $<9$) of AMD. We used eye-level data to develop the prediction model. Because our objective is prediction rather than inference, we did not explicitly model within-subject (between-eye) correlation, consistent with prior work \cite{sun2020genome,zeng2025tdcoxsnn}. Baseline covariates used in the model are age, sex, education level ($\ge$ high school or not), smoking status (never, former, or current), and baseline AMD severity score. We also incorporated genetic information into the prediction models using two alternative approaches. In one approach, we included a genetic risk score (GRS) derived from 34 established AMD risk SNPs, as previously calculated in \cite{ding2017bivariate}. Alternatively, we included individual SNPs identified from a prior GWAS, selecting the top variants at significance thresholds of 
$p = 10^{-7}$, $10^{-6}$, $10^{-5}$, and $10^{-4}$. The corresponding numbers of SNPs included at these significance thresholds were 89, 162, 663, and 1497. The characteristics of the analysis dataset are summarized in Table \ref{table: descriptive areds}. 

\begin{table}[ht]
\centering
\begin{threeparttable}
\caption{Baseline characteristics of analysis samples in AREDS and AREDS2 datasets}
\label{table: descriptive areds}
\begin{tabular}{lc}
\toprule
\textbf{Variable} & N=7803 \tnote{a} \\
\midrule
\textbf{Age (Years)} & \\
Mean (SD) & 69.5 (6.2) \\  
\midrule
\textbf{Gender (n, \%)} & \\
Female & 4466 (57\%) \\
Male & 3337 (43\%) \\
\midrule
\textbf{Education (n, \%)} & \\
$\leq$High School & 2369 (30\%) \\
$>$High School & 5434 (70\%)\\
\midrule
\textbf{Smoking Status (n, \%)} & \\
Never & 3623 (46 \%) \\
Former & 3752 (48\%) \\
Current & 428 (6 \%) \\
\midrule
\textbf{Baseline Severity Score} & \\
Mean (SD) & 4.2 (2.5) \\
\midrule
\textbf{Genetic Risk Score (GRS)} \tnote{b} & \\
Mean (SD) & 1.0 (0.14) \\
\midrule
\textbf{Censoring Type (n, \%)} & \\
Interval-censored & 2080 (27\%)\\
Right-censored & 5723 (73\%)\\ 
\midrule
\textbf{Study Phase (n, \%)} & \\
AREDS 1 & 5015 (64\%)\\
AREDS 2 & 2788 (36\%)\\ 
\bottomrule
\end{tabular}
\begin{tablenotes}[flushleft]\footnotesize
    \item \tnote{a} All numbers in the table were calculated by eye.
    \item \tnote{b} The GRS was calculated from a previous article \cite{ding2017bivariate}.
\end{tablenotes}
\end{threeparttable}
\end{table}

\subsubsection{Prediction Model Comparisons}
\label{subsubsec:prediction-AREDS}

We developed several models to predict time-to-late AMD (defined as a severity score $\geq 9$), using interval-censored data from AREDS and AREDS2, incorporating both genetic and non-genetic risk factors such as age, education level, and smoking status. We excluded the baseline severity score from the primary models to avoid a ceiling effect, as its strong predictive power could mask differences in model performance. We compared our ICODEN model with three other Cox-based models, including a model with only non-genetic factors (``Non-genetic''), a model with the GRS and non-genetic factors (``GRS''), and a model with all individual SNPs and non-genetic factors (``IcenReg''). All three comparison models were fitted using the package \texttt{IcenReg}. 

The model comparison results are summarized in Table \ref{table: AREDS model compare}. The non-genetic model shows the poorest performance in terms of IBS and  $d_{out}$, although it achieves the best performance in $p_{out}$, which is attributed to the high right-censoring rate of over $70 \%$ in the dataset. The GRS model performs similarly to the IcenReg model with 89 or 162 SNPs. In contrast, our proposed ICODEN model outperforms all other methods in IBS and $d_{out}$, particularly when a large number of SNPs (663 and 1497) are included. Consistent with the simulation and ADNI data analyses, the IcenReg model failed to handle datasets with such large numbers of SNPs.

\begin{table}[ht]
\centering
\begin{threeparttable}
\caption{Performance comparison of various models using AREDS and AREDS2 data. Values are reported as mean (SD). ICODEN denotes the proposed method; non-genetic, GRS, and IcenReg are Cox-PH models that include only the non-genetic factor, additionally the GRS score, and all SNPs, respectively.}
\label{table: AREDS model compare}
\begin{tabular}{llllllll}
\toprule
& P-Value\tnote{a} & SNPs & Non-genetic\tnote{b} & GRS\tnote{b} & IcenReg & ICODEN \\ 
\midrule

IBS& 
$<1\times 10^{-7}$ & 89 & 0.122 (0.004) & 0.110 (0.006) & 0.113 (0.004) & 0.107 (0.005)\\
& $<1\times 10^{-6}$ & 162 & 0.122 (0.004) & 0.110 (0.006) & 0.111 (0.005) & 0.109 (0.005)\\
& $<1\times 10^{-5}$ & 663 & 0.122 (0.004) & 0.110 (0.006) & -- & 0.103 (0.006)\\ 
& $<1\times 10^{-4}$ & 1497 & 0.122 (0.004) & 0.110 (0.006) & -- & 0.101 (0.008) \\
\midrule

$p_{out}$& 
$<1\times 10^{-7}$ & 89 & 0.284 (0.014) & 0.301 (0.013) & 0.296 (0.014)& 0.308 (0.018)\\
& $<1\times 10^{-6}$ & 162 & 0.284 (0.014) & 0.301 (0.013) & 0.305 (0.019) & 0.303 (0.022)\\
& $<1\times 10^{-5}$ & 663 & 0.284 (0.014) & 0.301 (0.013) & -- & 0.322 (0.016)\\ 
& $<1\times 10^{-4}$ & 1497 & 0.284 (0.014) & 0.301 (0.013) & -- & 0.317 (0.023) \\
\midrule

$d_{out}$& 
$<1\times 10^{-7}$ & 89 & 1.89 (0.12) & 1.68 (0.12) & 1.88  (0.09) & 1.69 (0.12)\\
& $<1\times 10^{-6}$ & 162 & 1.89 (0.12) & 1.68 (0.12) & 1.70 (0.12) & 1.75 (0.14)\\
& $<1\times 10^{-5}$ & 663 & 1.89 (0.12) & 1.68 (0.12) & -- & 1.56 (0.12)\\ 
& $<1\times 10^{-4}$ & 1497 & 1.89 (0.12) & 1.68 (0.12) & -- & 1.51 (0.14)\\
\bottomrule
\end{tabular}
\begin{tablenotes}[flushleft]\footnotesize
    \item \tnote{a} GWAS P-value threshold from a previous study \cite{sun2020genome}.
    \item \tnote{b} The non-genetic and GRS models are invariant to the choice of p-value thresholds.
\end{tablenotes}
\end{threeparttable}
\end{table}

\subsubsection{Subgroup Identification}
\label{subsubsec:subgroup-AREDS}

\begin{figure}[ht]
    \centering
    \includegraphics[width=0.7\textwidth]{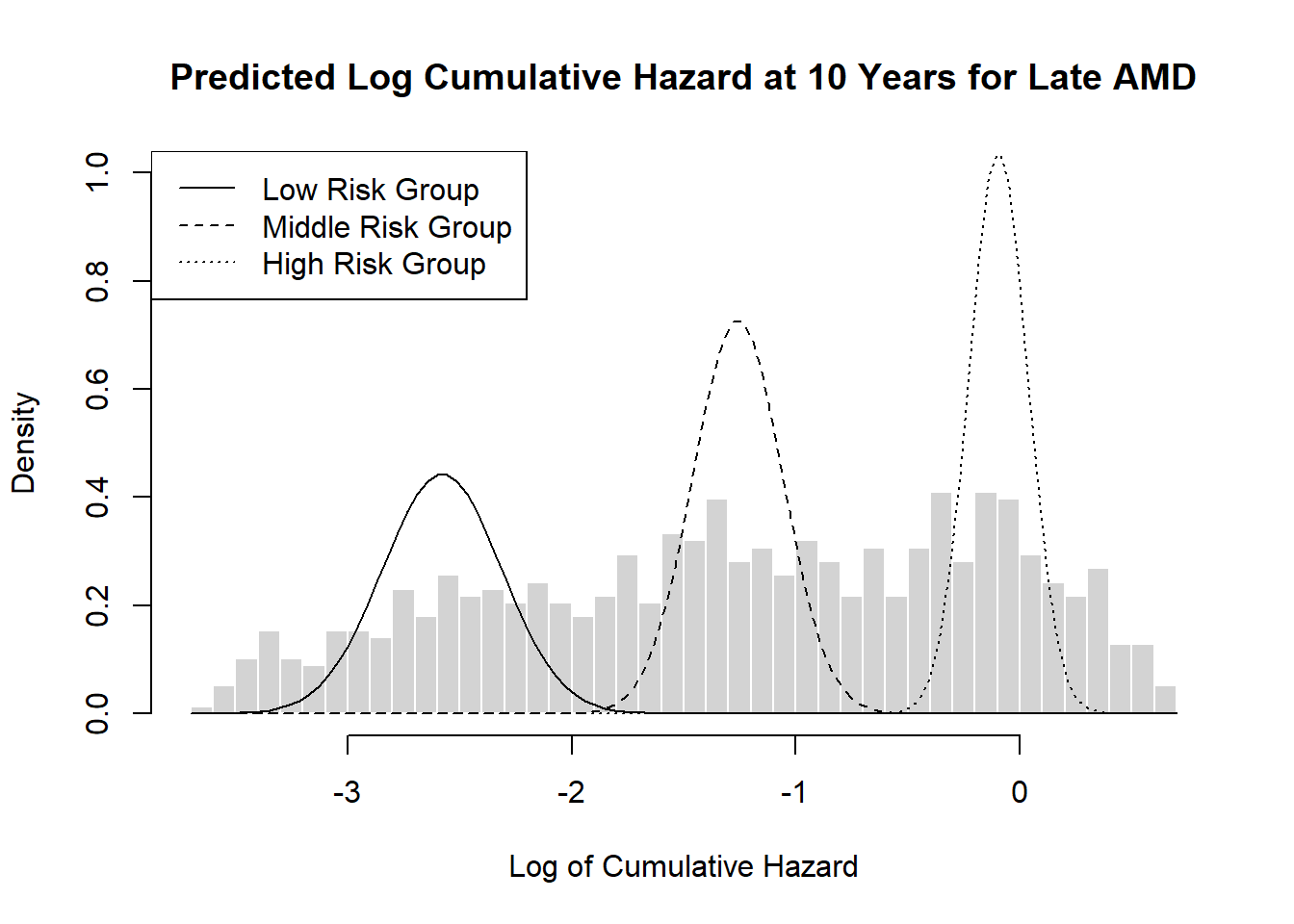}
    \includegraphics[width=0.7\textwidth]{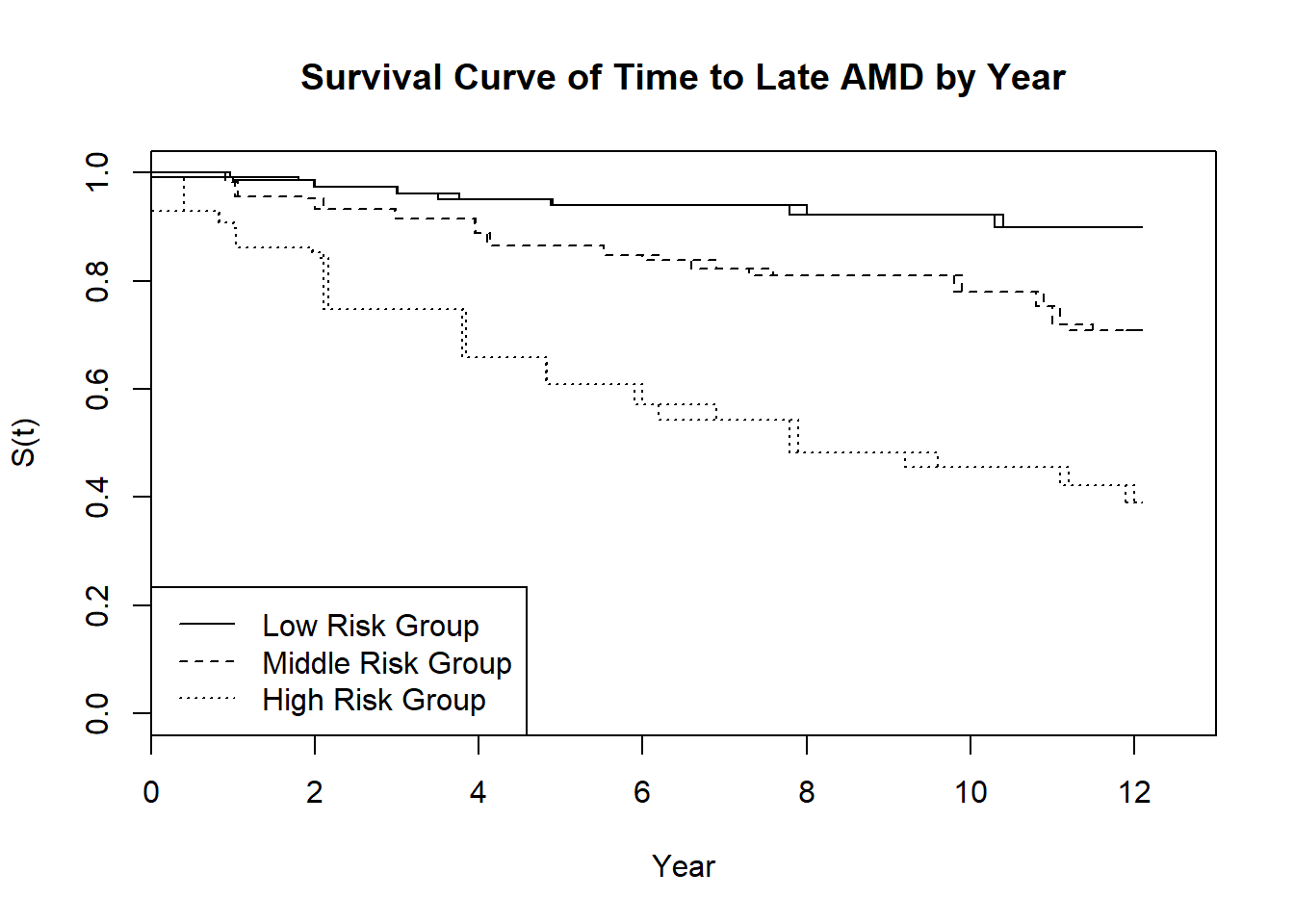}
    \caption{Gaussian mixture model-based subgrouping in AREDS and AREDS2 Data. The left panel shows the density distribution of predicted log cumulative hazard at 10 years, with the fitted Gaussian components corresponding to three different risk groups. The right panel shows the Turnbull survival estimates for time-to-late AMD in different risk subgroups.}
    \label{AREDS gaussian mixture model}
\end{figure}

After fitting the model with 89 SNP predictors on the training dataset (90\% of the data), we identified subgroups in the hold-out test data (10\% of the data) based on the predicted 10-year cumulative hazards. Using a Gaussian mixture model, we identified three subgroups with relatively balanced sizes. The left panel of Figure \ref{AREDS gaussian mixture model} shows the distribution of the estimated 10-year cumulative hazards in the test data.

As shown in the right panel of Figure \ref{AREDS gaussian mixture model}, the Turnbull-estimated survival functions for the three subgroups differ substantially, as confirmed by the log-rank test ($P<2.2\times10^{-16}$). This demonstrates that the prediction model developed by ICODEN can successfully identify subgroups with distinct risk profiles. The GRS also differs significantly across three subgroups (see Supplementary Material Section S4). This finding is consistent with previous work showing that GRS is a strong predictor of AMD progression \cite{ding2017bivariate}. 

\section{Conclusion}
\label{section:conclusion/discussion}

In this work, we developed a flexible method that integrates ODE with NN to analyze interval-censored survival data. Unlike traditional approaches for interval-censored outcomes, our framework relies on mild modeling assumptions and allows non-proportional hazards. By using NN, the method can capture nonlinear covariate effects and effectively handle high-dimensional covariates.

Both simulation studies and real-world applications demonstrate the strong performance of our method in predicting disease progression from interval-censored data. In real-world applications to AD and AMD studies, our method successfully identified meaningful patient subgroups with distinct progression risk profiles. The key differentiating features between these subgroups align well with findings from previous research.

Extending ICODEN to accommodate competing risks is an interesting direction for future research, as competing risks are common in clinical and epidemiological studies with survival outcomes. In such settings, one may consider modeling either a cause-specific hazard or a cumulative incidence function through NN under the ODE framework. This would provide a flexible modeling framework for competing risks data under interval censoring. In addition, the interpretability of ICODEN-based prediction models can be further improved by using advanced explanation techniques, such as local interpretable model-agnostic explanations, which provide more detailed and interpretable insights. We leave these extension works to interested readers and researchers.

\subsection*{Supporting information}
Supplemental material for this article is available online. The key functions for running simulations and real data analysis will be available on GitHub once the manuscript is accepted.

\subsection*{Conflict of interest}
The authors declare that they have no conflict of interest.

\subsection*{Data Availability Statement}
Data from all phases of ADNI are available at the online repository of \href{https://adni.loni.usc.edu/}{ADNI}. The data of AREDS and AREDS2 are available from the online repository dbGap (accession: phs000001.v3.p1 and phs001039.v1.p1, respectively).

\bibliographystyle{plain}
\bibliography{reference}

\end{document}


\maketitle

\section{Neural Network Training in the Simple Simulation Case}

\begin{figure}[H]
    \centering
    \includegraphics[width=0.6\textwidth]{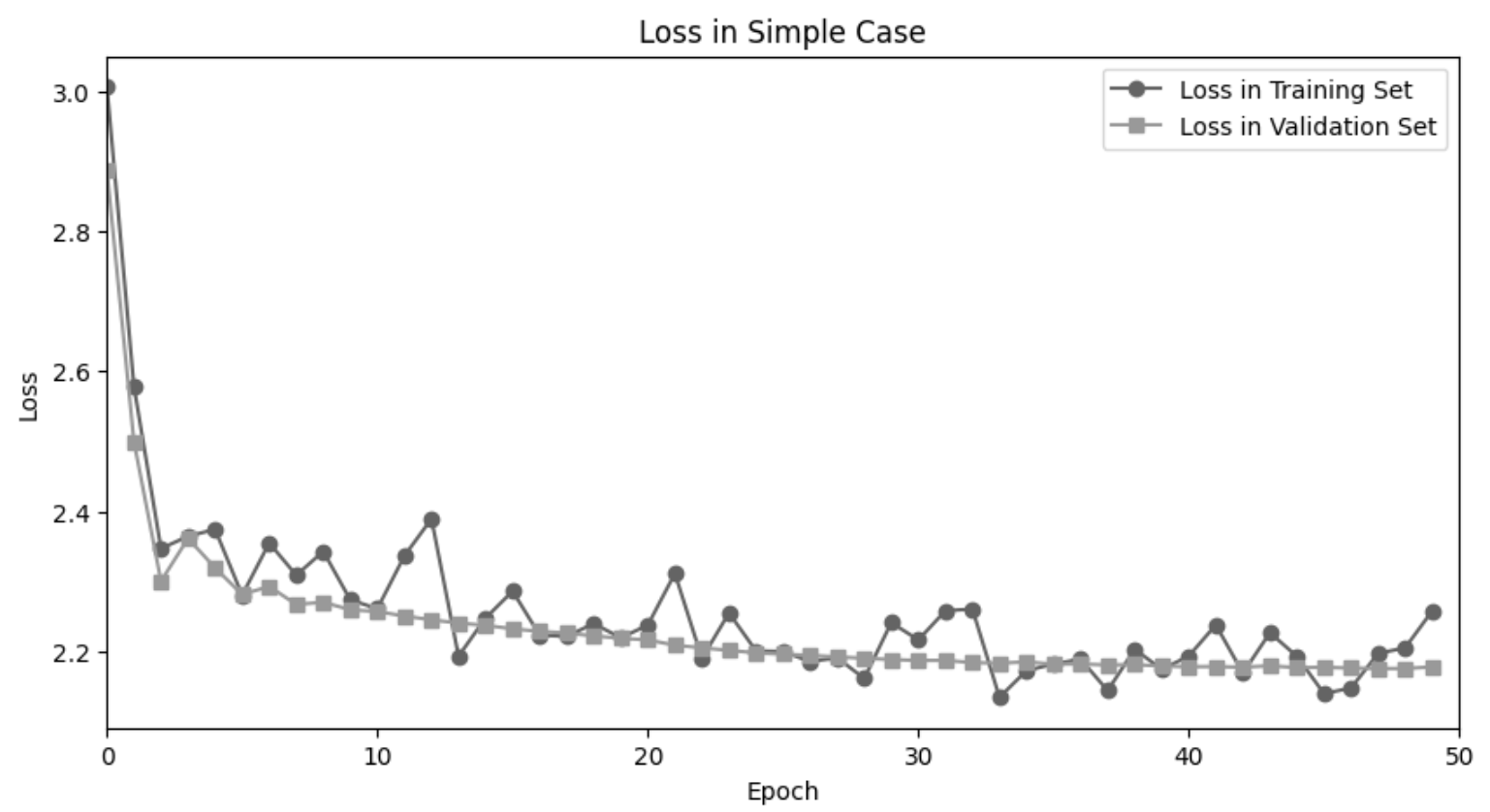}
    \caption{Loss function value by epoch in the training set and validation set in the simple simulation case. The epoch range is from 0 (the beginning of training) to epoch 50.}
    \label{simulation-example result}
\end{figure}

We examined the NN training by visualizing the loss across training epochs in this simple example, as illustrated in Figure \ref{simulation-example result}. The graph shows a clear and steady decline in validation loss, which eventually reaches a plateau at epoch about 20, indicating convergence of the model. The validation loss stabilizes after epoch 20, indicating that the model achieves stable performance beyond this point. 
This suggests that including the regularization term helps prevent overfitting, indicating that training for a sufficiently large number of epochs can yield stable performance.

\section{Hyperparameter Tuning in Simulation and Real Data Analysis}

\begin{figure}[H]
    \centering
    \includegraphics[width=0.6\textwidth]{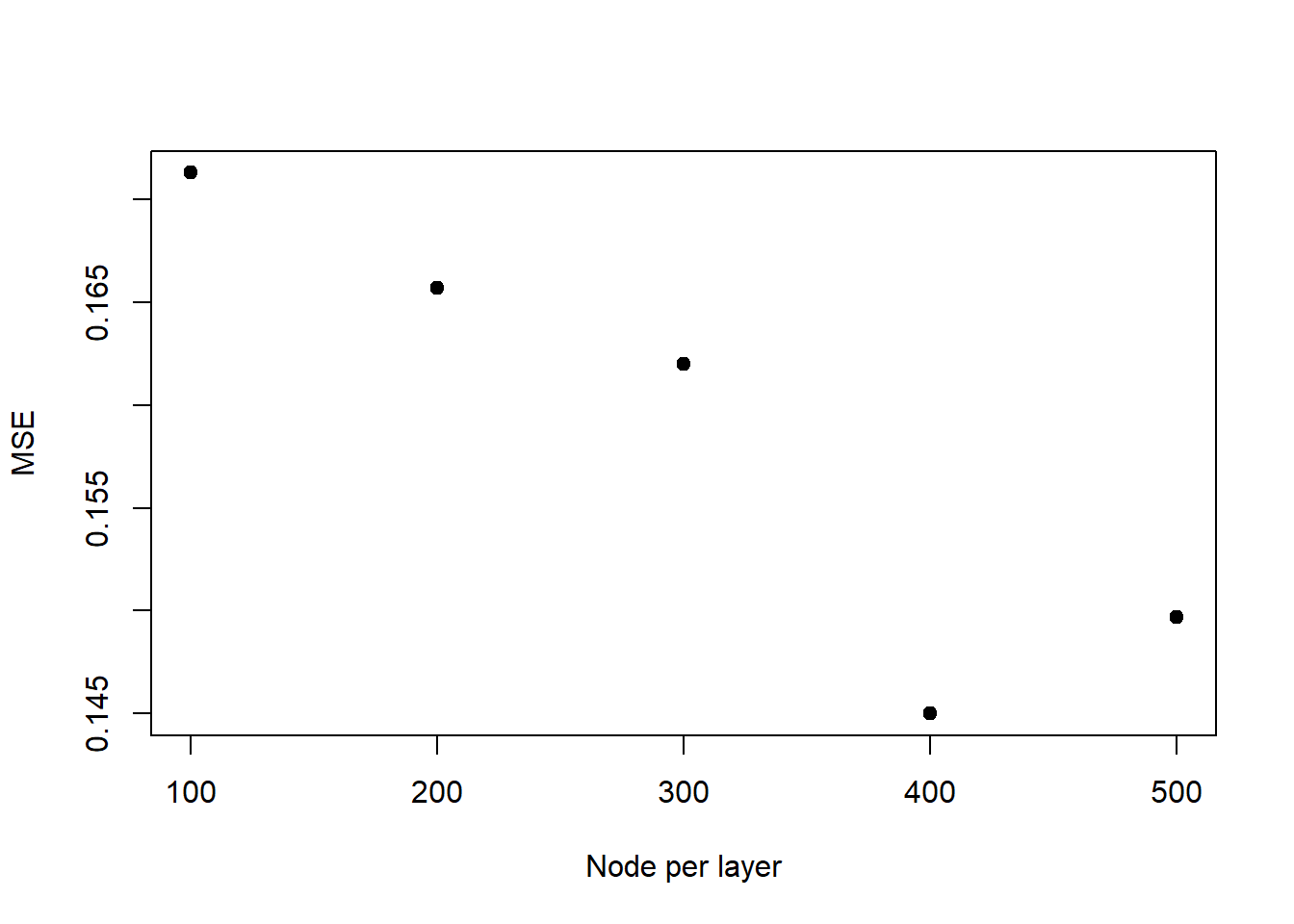}
    \includegraphics[width=0.6\textwidth]{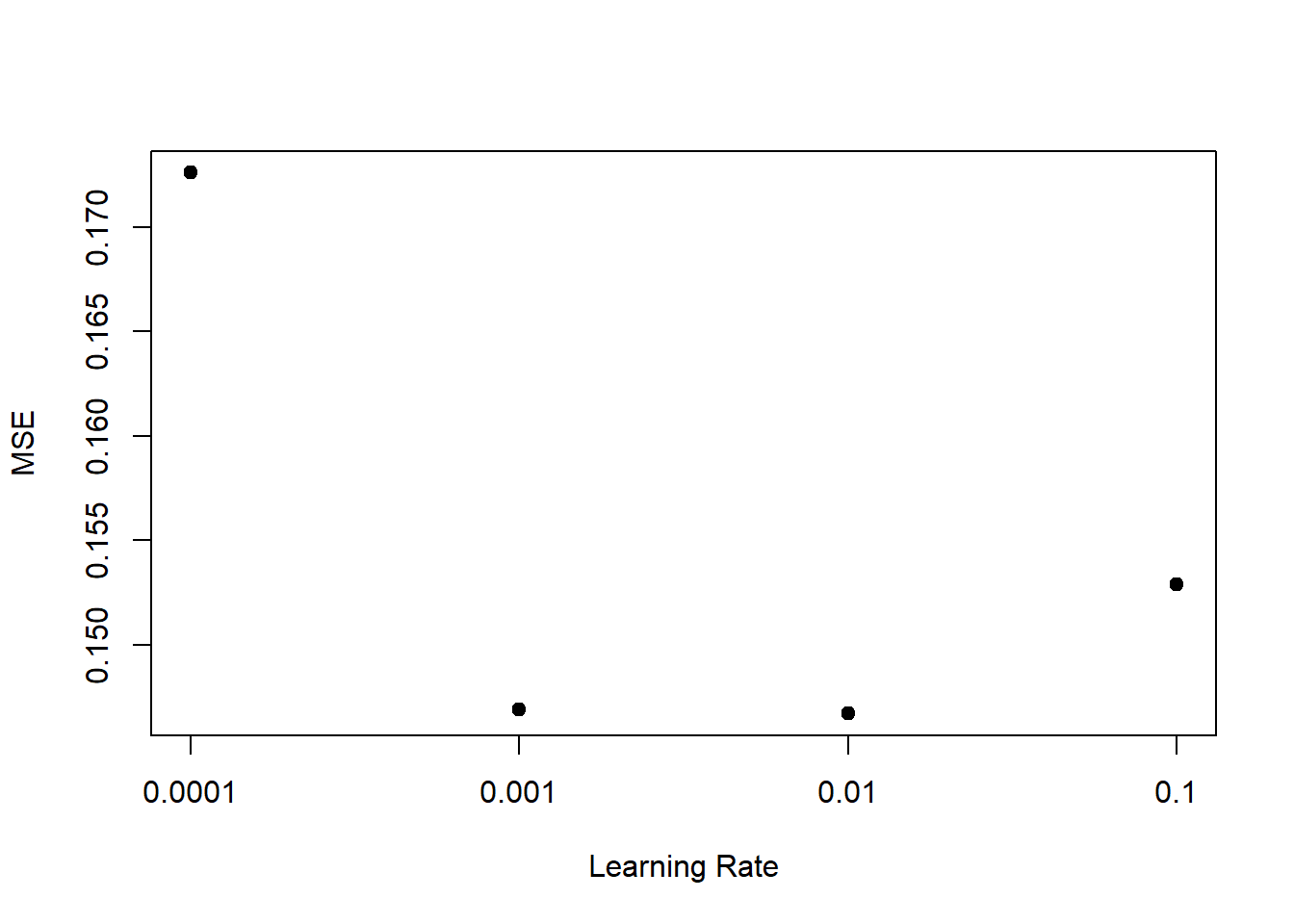}
    \includegraphics[width=0.6\textwidth]{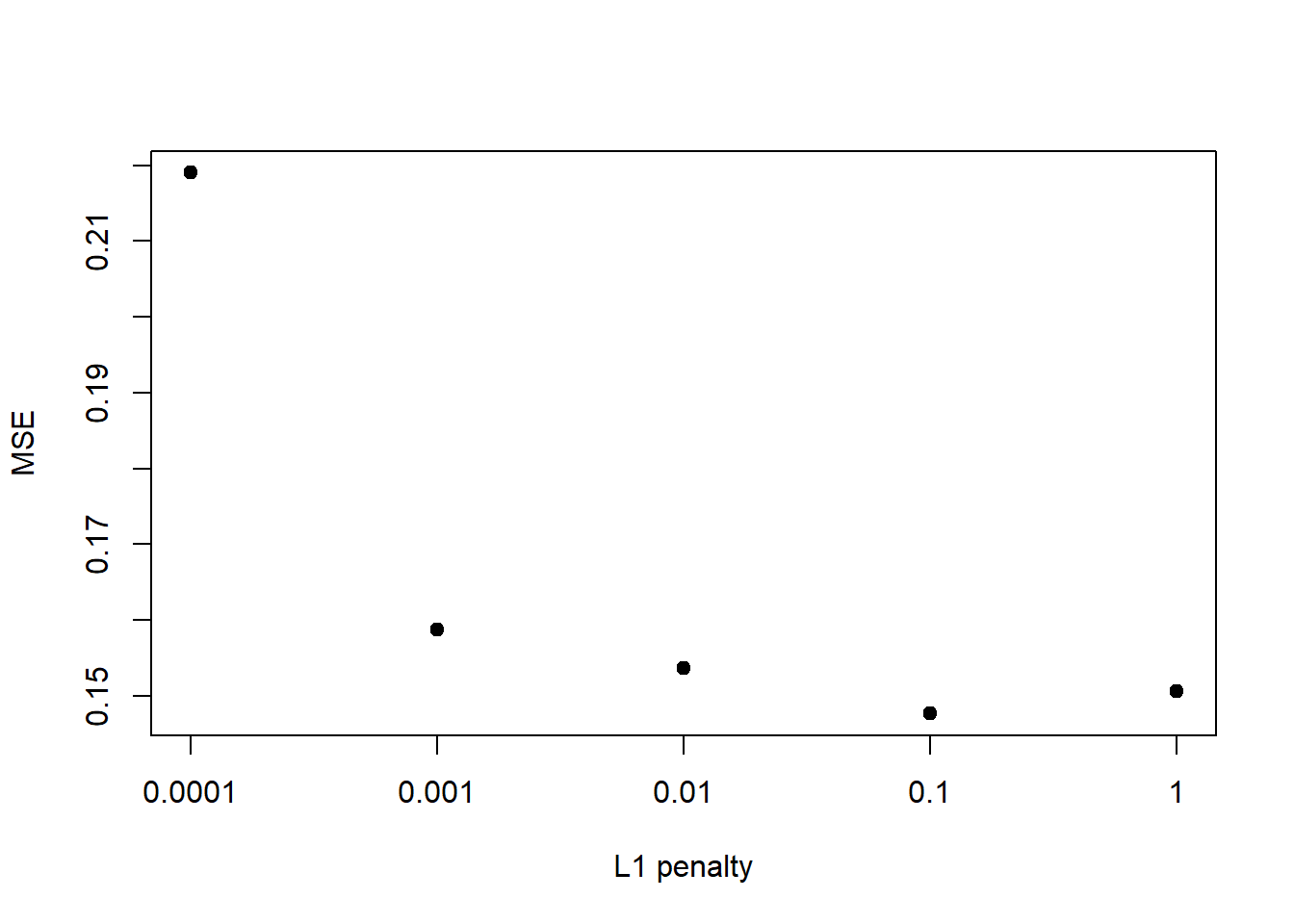}
    \caption{MSE in validation set through OAT hyperparameter tuning in synthetic simulation. The first graph shows the change in MSE for different numbers of nodes per layer. The second graph shows the MSE under different learning rates. The third graph shows the MSE with different choices of $L_1$ penalty parameter.}
    \label{tuning-synthetic}
\end{figure}

\begin{figure}[H]
    \centering
    \includegraphics[width=0.6\textwidth]{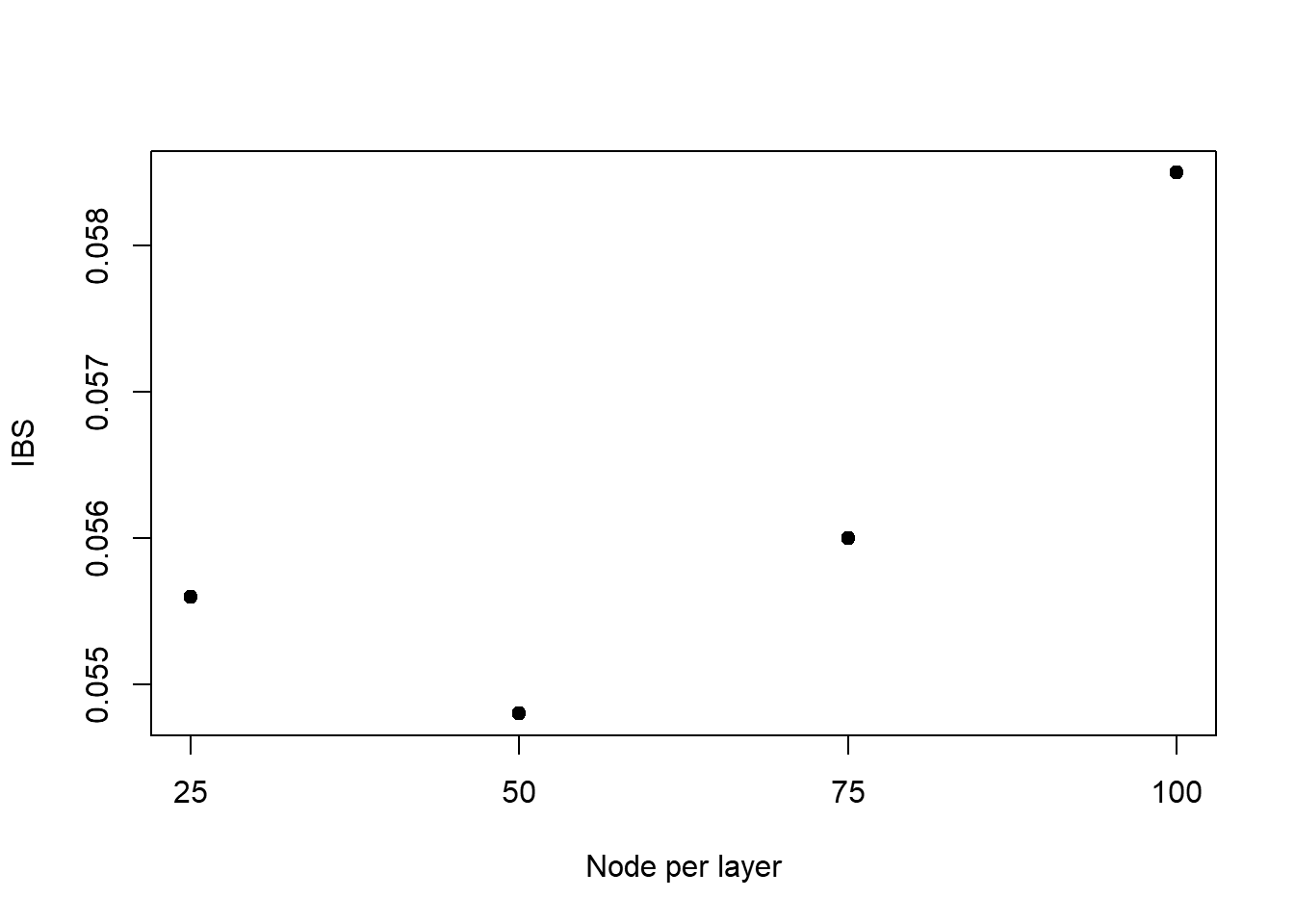}
    \includegraphics[width=0.6\textwidth]{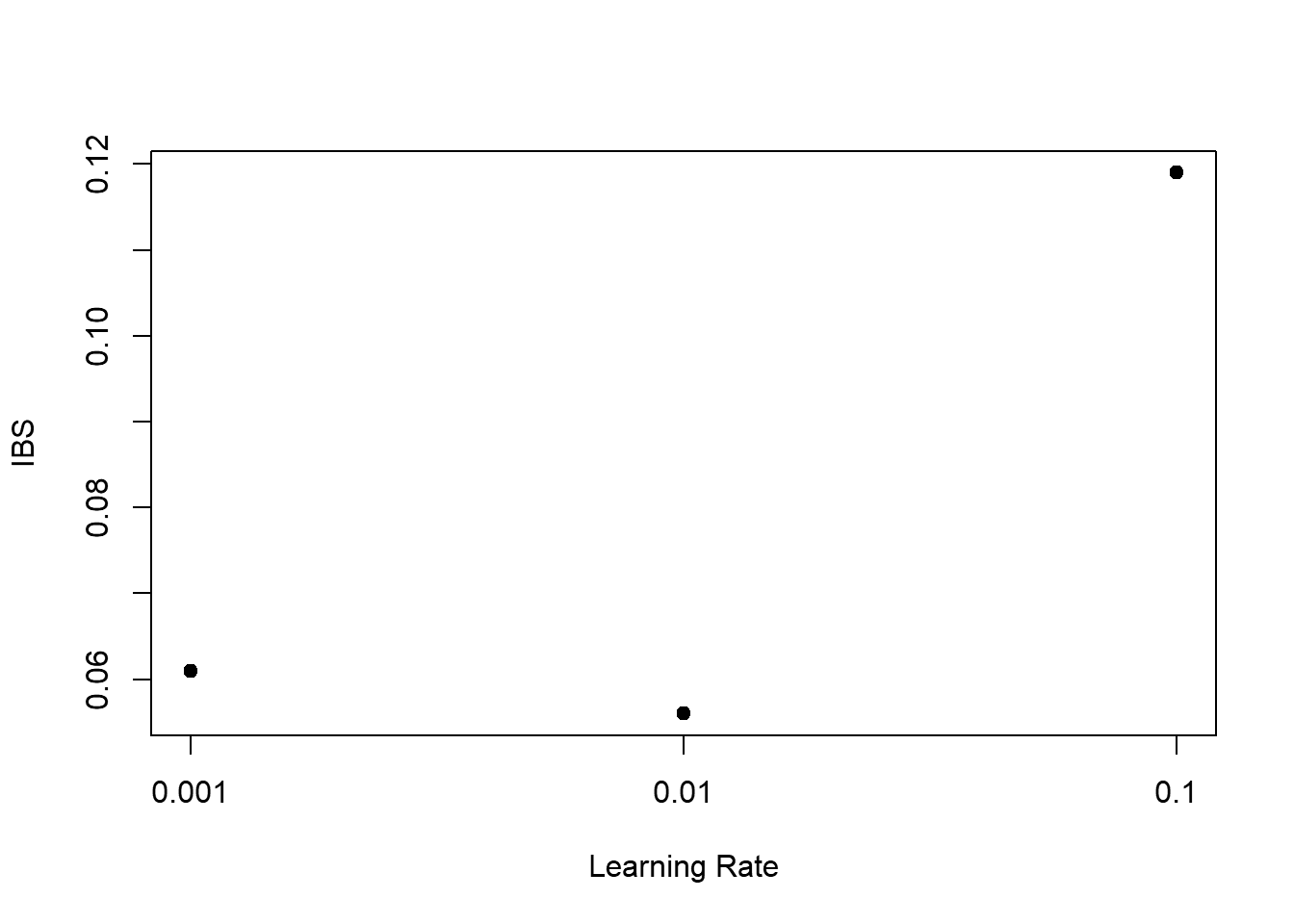}
    \includegraphics[width=0.6\textwidth]{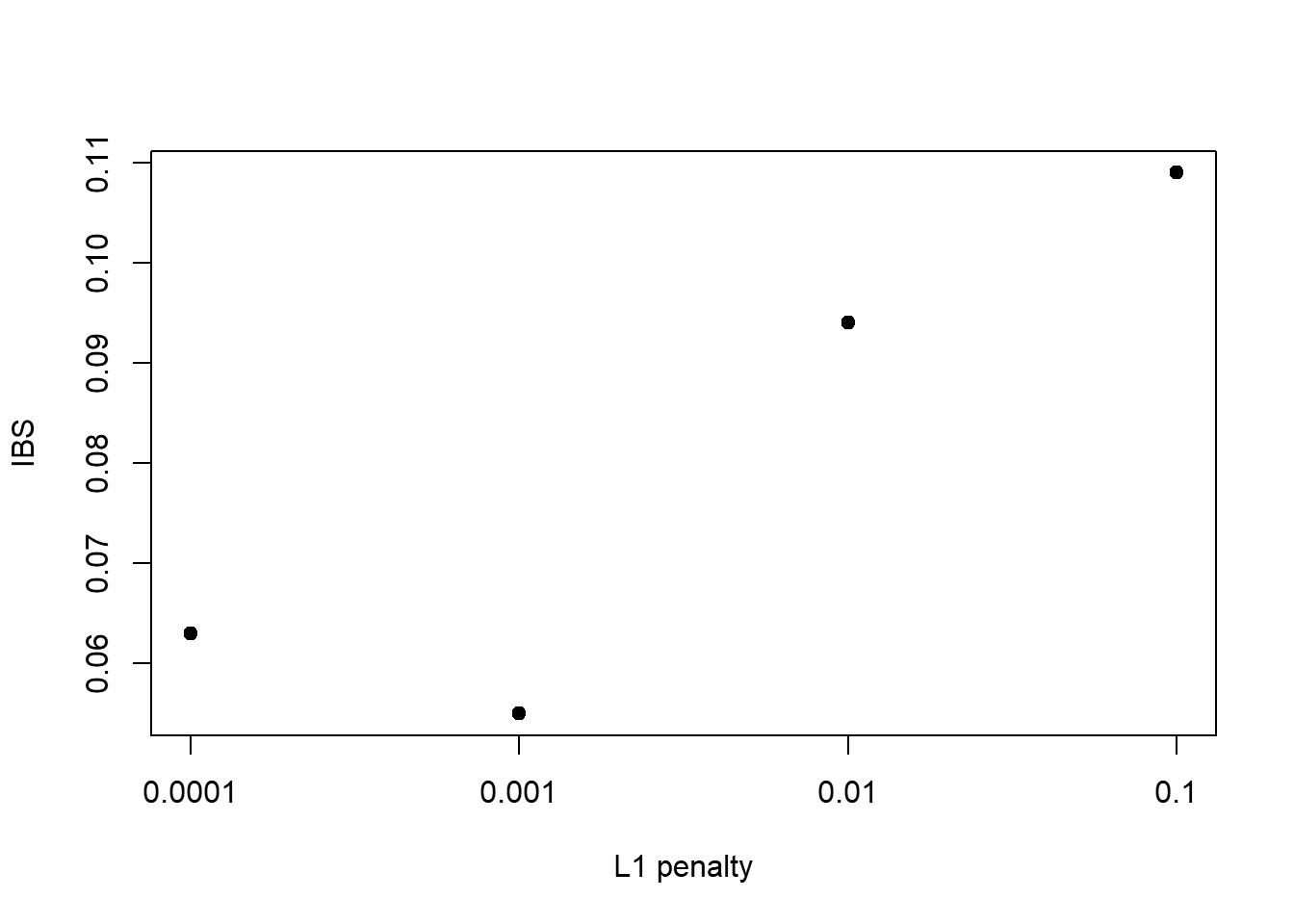}
    \caption{IBS in validation set through OAT hyperparameter tuning in ADNI with 71 SNPs. The first graph shows the IBS score change for different numbers of nodes per layer. The second graph shows the IBS score under different learning rates. The third graph shows the IBS score with different choices of $L_1$ penalty parameters.}
    \label{tuning-ADNI}
\end{figure}

We explored various hyperparameters in the ICODEN model (Table \ref{table: hyperparameter values}), and some hyperparameters used in specific scenarios are listed below (Table \ref{table: hyperparameter used}). We used a one-at-a-time (OAT) hyperparameter tuning approach: for each hyperparameter, we held all the others fixed at their values and then varied the target hyperparameter over the predefined grid. We repeated this procedure for every hyperparameter and selected the settings that maximized performance. Figures \ref{tuning-synthetic} and \ref{tuning-ADNI} present the evaluation metrics across tuning parameters for the synthetic simulation and the ADNI dataset with 71 SNPs. 

\begin{table}[H]
\centering
\caption{Hyperparameter values considered during tuning}
\label{table: hyperparameter values}
\begin{tabular}{ll}
\toprule
    Hyperparameter & Values Considered \\
\midrule
    Hidden Layers & 2 \\
    Nodes per Layer & From 10 to 500\\
    $L_1$ Parameter & 1, 0.1, 0.01, 0.001, 0.0001 \\
    Batch Size & 100, 400 \\
    Epoch Size & 50, 100 \\
    Learning Rate & 0.1, 0.01, 0.001, 0.0001 \\
\bottomrule
\end{tabular}
\end{table}

\begin{table}[H]
\centering
\caption{Hyperparameters used in simulation and real data analysis}
\begin{threeparttable}
\label{table: hyperparameter used}
\begin{tabular}{lllll}
\toprule
                   & Simple Example    & Simulations\tnote{a} & ADNI\tnote{b} & AREDS\tnote{c} \\
\midrule
    Hidden Layers & 2 & 2 & 2 & 2 \\
    Nodes per Layer & 10 & 400 & 50 & 100\\
    $L_1$ Parameter & 0.01 & 0.1 & 0.001 & 0.001\\
    Batch Size & 100 & 400 & 400 & 400\\
    Epoch Size & 50 & 100 & 100 & 100\\
    Learning Rate & 0.1 & 0.001 & 0.01 & 0.01 \\
\bottomrule
\end{tabular}
\begin{tablenotes}[flushleft]\footnotesize
    \item \tnote{a} For simulation Scenario 4 with $p=300$ and synthetic simulation. 
    \item \tnote{b} For the 71 SNPs case. 
    \item \tnote{c} For the 89 SNPs case. 
\end{tablenotes}
\end{threeparttable}
\end{table}

\section{Time-Rescaling Trick to Accelerate the Computation}

Every individual has a different predictor $X_i$, so the ordinary differential equation needs to be solved for each individual. After solving the ODE for every individual, one can compute the gradient and update the parameter $\theta$. There are two major steps in the NN computation: forward pass and backward pass. In the forward pass step, the feature $x$ is passed through hidden layers to obtain the prediction. Although there may be no analytical form of solution, we can compute $\Lambda$ numerically with the ODE solver at any given time $t$, with the network structure $f$ and the initial value at time $0$. In the backward pass, the key part is back-propagation to compute gradients of the log-likelihood. We use adjoint sensitivity analysis as described in \cite{DUMMY:2(SODEN)} to compute the gradient using augmented ODE from \cite{DUMMY:2(SODEN)}. Given observed time $t_i$, let $b(t)$ be an adjoint function such that 
\begin{gather*} 
\begin{cases} 
b'(t) = -\frac{\partial f_{\theta}}{\partial \Lambda}b(t) \\ 
b(t_i) = 1 
\end{cases}, 
\end{gather*} 
then the gradient is calculated from solving the following augmented ODE 
\begin{gather*} 
v'(t) = \bigg{[}f_{\theta}(\Lambda(t|x_i,\theta),t;x_i),-b(t)\frac{\partial f_{\theta}}{\partial \Lambda}, -b(t)\frac{\partial f_{\theta}}{\partial \theta} \bigg{]}, \\ 
v(t_i) = [\Lambda(t_i|x_i,\theta),1,\underbrace{0,\ldots,0}_{\text{length of} \ \theta}]. 
\end{gather*}

\section{Additional Subgroup Analysis Results in Real Data Analysis}

\begin{figure}[H]
    \centering
    \includegraphics[width=0.8\textwidth]{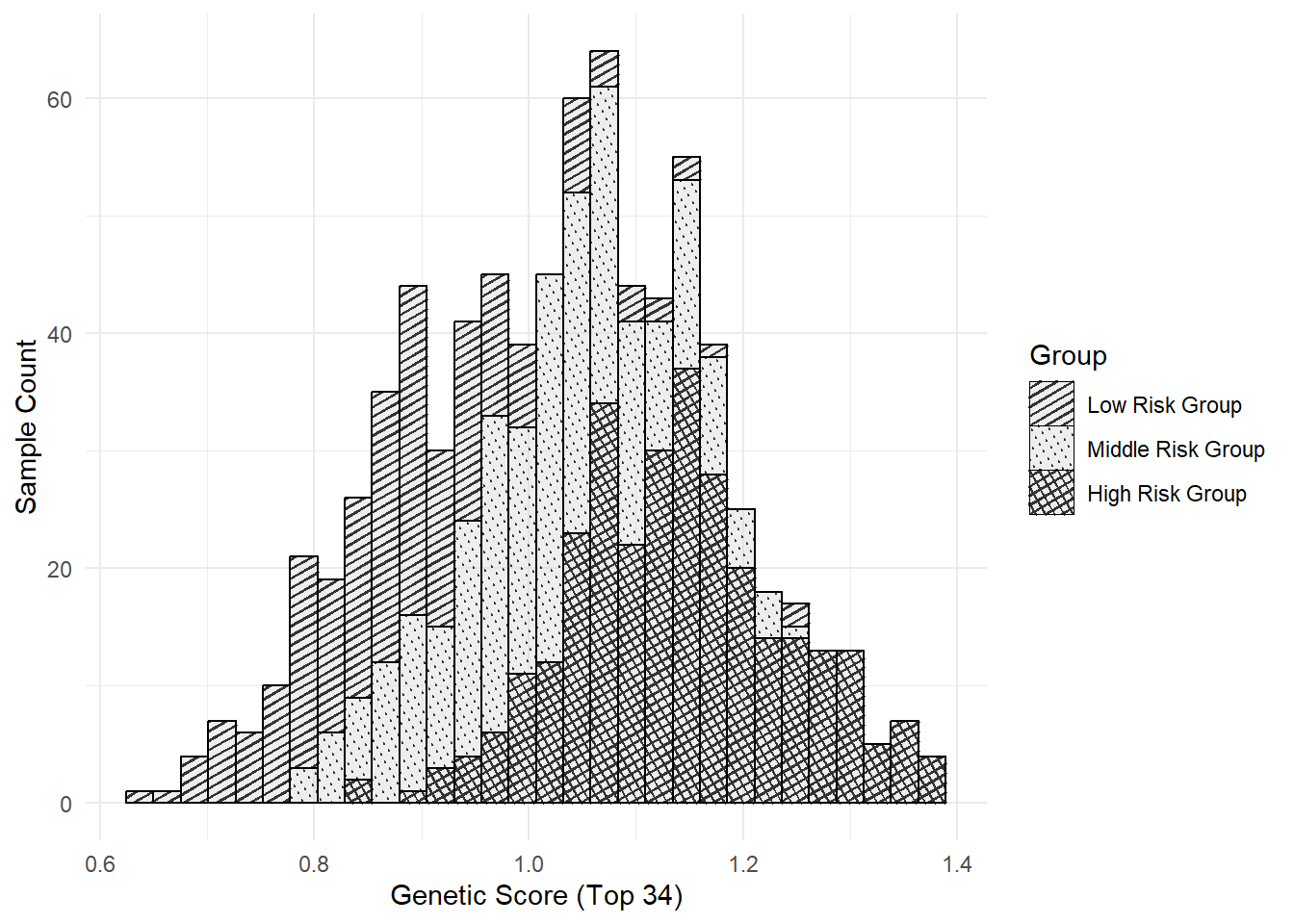}
    \caption{Histogram of genetic risk scores across three risk subgroups in the AREDS \& AREDS2 datasets.}
    \label{AREDS grs by subgroup}
\end{figure}

The results of the subgroup analysis are summarized below. Table \ref{SNPs between groups} reports the top 10 SNPs that show the strongest difference (smallest $p$-value) between the two identified ADNI subgroups. Figure \ref{AREDS grs by subgroup} illustrates the distribution of the genetic risk score among the three identified subgroups of AREDS \& AREDS2, with the differences highly significant ($p < 2.2 \times 10^{-16}$).

\begin{table}[H]
    \centering
    \caption{Contingency table and P-value of the top 10 SNPs with the most significant difference between two risk subgroups in ADNI analysis.}
    \label{SNPs between groups}
    \resizebox{\textwidth}{!}{
    \begin{tabular}{llllllll}
    \toprule
        Chromosome & Gene & SNP & Allele by Group &0 & 1& 2& P-value \\ 
    \midrule
        19 & \textit{APOE} & $rs429358$ & High Risk & 481 & 534 & 161 & $4.82 \times 10^{-58}$ \\
           &      &            & Low Risk &  460 & 103 & 1 & \\ 
    \midrule
        1 & \textit{SORT1} & $rs4970843$& High Risk & 378  &579 & 219 & $1.73 \times 10^{-21}$ \\
          &       &            & Low Risk &   87  &265  & 212& \\ 
    \midrule
        3 & \textit{SRGAP3} & $rs147648392$  & High Risk & 1114 & 62 &0 &$1.98 \times 10^{-11}$ \\
          &        &                & Low Risk & 480& 80 & 4 & \\ 
    \midrule
        2 & \textit{MMADHC} & $rs1113138$ & High Risk & 1007&164 & 5 &$2.16 \times 10^{-09}$ \\
          &        &               & Low Risk & 422& 126 & 16 & \\ 
    \midrule
        9 & \textit{PHF24} & $rs17355465$  &  High Risk & 1090& 84 & 2&$4.40 \times 10^{-09}$ \\
           &      &              &  Low Risk &   468&93& 3& \\ 
    \midrule
        19 & \textit{TOMM40} & $rs741780$ & High Risk & 500 &525 &151&$ 6.20 \times 10^{-09}$ \\
           &        &           & Low Risk & 158 & 295& 111& \\ 
    \midrule
        19 & \textit{APOE} &$rs7412$ & High Risk & 1106 & 69  & 1& $2.19 \times 10^{-08}$ \\
          &      &          & Low Risk &   482& 80 & 2 & \\ 
    \midrule
        1 & \textit{UBR4} &$rs72650394$ & High Risk & 988 &183&5&$8.18 \times 10^{-08}$ \\
          &      &               &Low Risk &  529&34 &1 & \\ 
    \midrule
        1 &\textit{CAMTA} & $rs80099124$  & High Risk & 682 &418 & 76&$2.35 \times 10^{-07}$ \\
          &      &               &Low Risk &  400& 149&  15& \\ 
    \midrule
        8 &\textit{NRG1} & $rs79622257$ & High Risk & 709& 401&  66&$3.39 \times 10^{-07}$ \\
          &     &           &   Low Risk &  412& 139&  13& \\ 
    \bottomrule
    \end{tabular}
    }
\end{table}

\bibliographystyle{unsrt}  
\bibliography{reference}